\documentclass[lettersize,journal]{IEEEtran}

\usepackage{amsmath,amsfonts}
\usepackage{algorithmic}
\usepackage{algorithm}
\usepackage{array}
\usepackage[caption=false,font=normalsize,labelfont=sf,textfont=sf]{subfig}
\usepackage{textcomp}
\usepackage{stfloats}
\usepackage{url}
\usepackage{verbatim}
\usepackage{graphicx}
\usepackage{cite}
\usepackage[numbers]{natbib}
\usepackage[utf8]{inputenc} % allow utf-8 input
\usepackage[T1]{fontenc}    % use 8-bit T1 fonts
\usepackage{xcolor}
\definecolor{cvprblue}{rgb}{0.21, 0.49, 0.74}
\usepackage[colorlinks, citecolor=cvprblue, urlcolor=blue]{hyperref}       % hyperlinks
\usepackage{url}            % simple URL typesetting
\usepackage{booktabs}       % professional-quality tables
\usepackage{amsfonts}       % blackboard math symbols
\usepackage{nicefrac}       % compact symbols for 1/2, etc.
\usepackage{microtype}      % microtypography
\usepackage{xcolor}         % colors
\usepackage{amsmath}

\usepackage{multirow}
\usepackage{tablefootnote}
\usepackage{makecell}

\def\E{\mathbb{E}}

\usepackage{forest}
\usepackage{array}
\usepackage{tikz}

\definecolor{deepgreen}{HTML}{057311}

\begin{document}
\title{Flow Matching Meets Biology and Life Science: A Survey}

\author{Zihao Li$^{*1}$, Zhichen Zeng$^{*1}$, Xiao Lin$^{*1}$, Feihao Fang$^1$, Yanru Qu$^1$, Zhe Xu$^{1,2}$, Zhining Liu$^1$, \\ Xuying Ning$^1$, Tianxin Wei$^1$, Ge Liu$^{\diamond1,3}$, Hanghang Tong$^{\diamond1}$, Jingrui He$^{\diamond1}$
% \\ $^1$ University of Illinois Urbana-Champaign, Champaign, IL, USA $^2$ Meta, Menlo Park, CA, USA \\ $^3$DOE Center for Advanced Bioenergy and Bioproducts Innovation,
% UIUC, Champaign, IL, USA
\\ $^1$ University of Illinois Urbana-Champaign $^2$ Meta \\ $^3$DOE Center for Advanced Bioenergy and Bioproducts Innovation, UIUC
\\ \vspace{3mm} \small{$^*$Equal Contributions. Project Lead: Zihao Li $^{\diamond}$Primary Corresponding author: Jingrui He (\href{mailto: jingrui@illinois.edu}{jingrui@illinois.edu}). \\ Other Corresponding authors: Hanghang Tong (\href{mailto: htong@illinois.edu}{htong@illinois.edu}), Ge Liu (\href{mailto: geliu@illinois.edu}{geliu@illinois.edu}), Zihao Li (\href{mailto: zihaoli5@illinois.edu}{zihaoli5@illinois.edu})}

\vspace{3mm}
\textbf{Article Link: \url{https://www.nature.com/articles/s44387-025-00066-y}}
\vspace{-5mm}
}

\markboth{Nature Portfolio Journal Artificial Intelligence 2026}{}

\maketitle

\begin{abstract}
    Over the past decade, advances in generative modeling, such as generative adversarial networks, masked autoencoders, and diffusion models, have significantly transformed biological research and discovery, enabling breakthroughs in molecule design, protein generation, catalysis discovery, drug discovery, and beyond. At the same time, biological applications have served as valuable testbeds for evaluating the capabilities of generative models. Recently, flow matching has emerged as a powerful and efficient alternative to diffusion-based generative modeling, with growing interest in its application to problems in biology and life sciences. This paper presents the first comprehensive survey of recent developments in flow matching and its applications in biological domains. We begin by systematically reviewing the foundations and variants of flow matching, and then categorize its applications into three major areas: biological sequence modeling, molecule generation and design, and peptide and protein generation. For each, we provide an in-depth review of recent progress. We also summarize commonly used datasets and software tools, and conclude with a discussion of potential future directions. 
    The corresponding curated resources are available at \url{https://github.com/Violet24K/Awesome-Flow-Matching-Meets-Biology}.
\end{abstract}


\section*{Introduction}
\label{sec: introduction}

Flow Matching (FM) \cite{lipman2023flow} has recently emerged as a powerful paradigm for generative modeling, offering a flexible and scalable framework applicable across a wide range of domains, such as computer vision \cite{lipman2023flow, DBLP:conf/iclr/Jin00XXJZHSML25}, and natural language processing~\cite{DBLP:conf/eacl/HuWAMFOS24, gat2024discrete}.
By constructing a continuous probability trajectory between simple and complex distributions, FM provides an efficient and principled method to model high-dimensional, structured data. While FM has demonstrated strong performance in conventional generative tasks such as image, video, and language synthesis, its potential extends far beyond these domains. In particular, its ability to model diverse modalities while preserving structural and geometric constraints makes it especially well-suited for applications in biology and life sciences.

\begin{figure}[t]
    \centering
    \includegraphics[width=\linewidth]{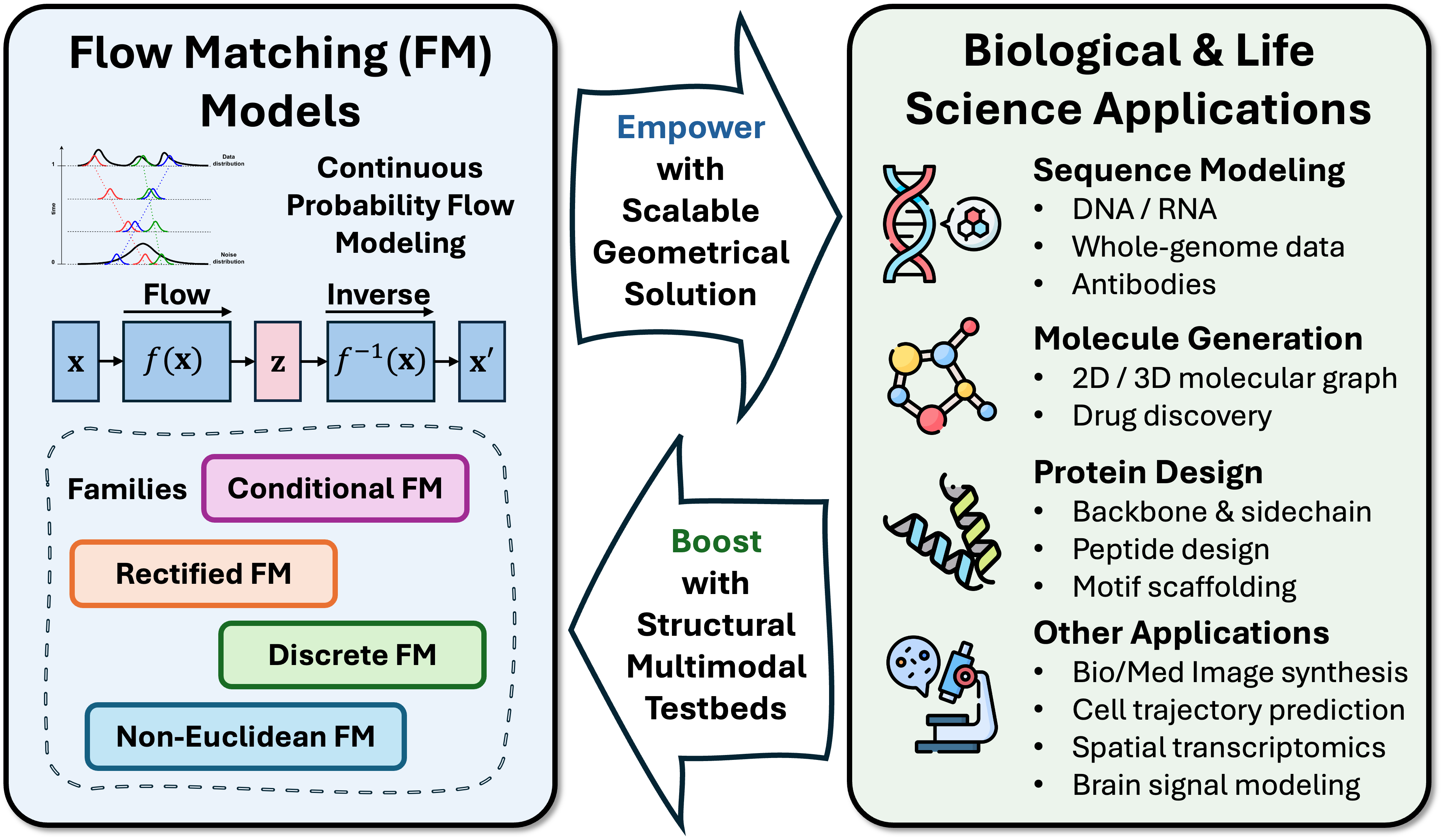}
    \vspace*{-15pt}
    \caption{
        Flow Matching Meets Biological and Life Sciences. Flow matching serves as a powerful generative modeling paradigm for a wide range of biological and life science applications. Conversely, these domains offer rich and diverse tasks for evaluating and advancing flow matching techniques. In this survey, we first present state-of-the-art flow matching models and their variants, then categorize their applications into four major areas: sequence modeling, molecule generation, protein design, and other emerging biological applications. The corresponding curated resources are available at \url{https://github.com/Violet24K/Awesome-Flow-Matching-Meets-Biology}.
    }
    \label{fig: intro}
    \vspace*{-10pt}
\end{figure}

At the same time, biological and life science applications present a natural testbed for FM (Figure \ref{fig: intro}). 
These tasks, ranging from genomic sequence modeling~\cite{church1984genomic, venter2001sequence, pareek2011sequencing}, molecular graph generation~\cite{luo20213d, sadybekov2023computational, mathur2017drug}, and protein structure prediction~\cite{abramson2024accurate, jumper2021highly, baek2023efficient}, to biomedical image synthesis~\cite{robb1999biomedical, tempany2001advances, webb2022introduction, rangayyan2004biomedical}, are often high-dimensional, multimodal, and governed by strict structural, physical, or biochemical constraints. 
In fact, they have already served as benchmarks for validating the performance of various generative modeling paradigms, such as Generative Adversarial Networks \cite{goodfellow2020generative, lan2020generative, lee2023recent}, Masked Autoencoders \cite{he2022masked, kraus2024masked, yuan2023proteinmae, chien2022maeeg}, and Diffusion Models \cite{ho2020denoising, guo2024diffusion, yang2023diffusion}.
Compared to traditional rule-based simulations~\cite{faeder2005rule, hwang2009rule, faeder2009rule, chylek2015modeling} and physics-driven models~\cite{willard2020integrating, newman2008physics, franklin2019introduction, baverstock2013life}, which often suffer from limited scalability and reliance on expert-crafted rules, these machine-learning-based generative models offer a data-driven alternative that can scale to complex biological systems, adapt to diverse modalities, and generalize beyond handcrafted constraints \cite{yelmen2023overview, anstine2023generative, bilodeau2022generative, xue2019advances, DBLP:conf/bigdataconf/FuH22, DBLP:journals/corr/abs-2412-21151, DBLP:journals/corr/abs-2504-07394, DBLP:conf/kdd/ZhengJLTH24, llmgraph}. By learning directly from empirical data, they enable the generation of biologically plausible outputs while significantly reducing the need for domain-specific assumptions.
FM, as a newer yet promising alternative, inherits key advantages from these models such as expressiveness, scalability, and data efficiency, while introducing a more stable training objective based on continuous probability flows. Its ability to generate high-quality samples with fewer inference steps makes it particularly appealing for biological applications, where modeling precision and computational efficiency are both critical.

\begin{figure*}[t]
\centering
    \includegraphics[width=0.9\textwidth]{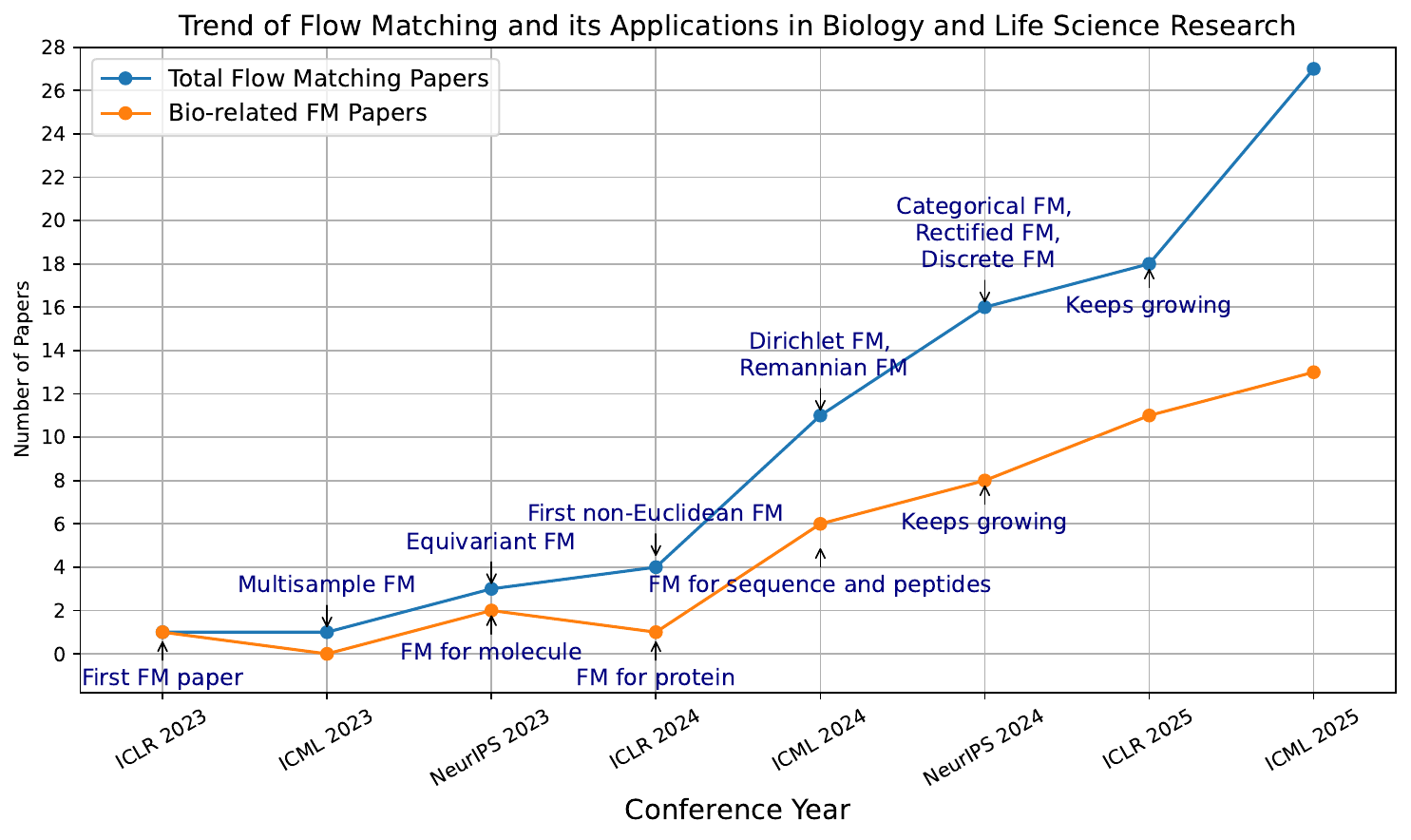}
    \caption{Trend of published papers on flow matching (FM) and its applications in biology and life sciences across major ML conferences from 2023 to 2025. The blue line indicates the total number of FM papers, while the orange line shows the subset focused on biological applications. Annotations highlight key milestones in FM and its adoption for molecule, sequence, and protein generation, illustrating the rapid growth and expanding interest in this area.}
    \label{fig: trend}
\end{figure*}

Interest in applying FM to biological problems is growing rapidly. As illustrated in Figure~\ref{fig: trend}, we have observed a steadily growing trend in the number of FM-related publications, with a visible rise in bio-related applications. 
The first biological applications appeared at NeurIPS 2023~\cite{DBLP:conf/nips/SongGXCLEZM23, klein2023equivariant}, both focusing on molecule generation. This momentum continued with the introduction of FM-based protein generation models at ICLR 2024~\cite{DBLP:conf/iclr/BoseAHFRLNKB024}, followed by further progress in biological sequence and peptide generation at ICML 2024. Beyond these milestones, 2024 and 2025 have seen the emergence of increasingly specialized FM variants, such as categorical FM \cite{DBLP:conf/nips/ChengL0L24}, rectified FM \cite{kornilov2024optimal}, and non-Euclidean formulations including Riemannian \cite{chen2024flow} and Dirichlet \cite{stark2024dirichlet} FM. 
Many of these have begun to find applications in structural biology, molecular conformation modeling, and biomedical imaging. 
More recently, NeurIPS 2025 features over 30 accepted FM papers, and ICLR 2026 received more than 150 FM-related submissions. As of the time this survey is under peer review (Nov 2025), these venues collectively include over 20 new FM-for-biology works. Since their proceedings are not yet public, we only cover the NeurIPS 2025 papers with available preprints and leave full coverage of these emerging results to future iterations.
This upward trajectory highlights not only the methodological innovation within FM, but also its growing relevance in life science domains that demand high-dimensional, structure-aware generative modeling.

As both FM and its biological applications evolve, the landscape has become increasingly fragmented, making it difficult to keep track of key developments and emerging trends. This survey addresses this gap by providing the first comprehensive review of FM in the context of biology and life sciences. We begin with a systematic overview of FM methods and variants, and then categorize their biological applications into three core areas: biological sequence modeling, molecule generation and design, and protein generation. We also review auxiliary topics such as bioimage modeling and spatial transcriptomics, summarize commonly used datasets and tools, and conclude with open challenges and future directions. Our goal is to offer an accessible entry point for newcomers, while equipping experienced researchers with a clear map of the field's current trajectory. Our curated resources are publicly available at \url{https://github.com/Violet24K/Awesome-Flow-Matching-Meets-Biology}.

\begin{figure*}[t]
	\centering
	\resizebox{\textwidth}{!}{
	\begin{forest}
  for tree={
  grow=east,
  reversed=true,
  anchor=base west,
  parent anchor=east,
  child anchor=west,
  base=left,
  font=\small,
  rectangle,
  draw,
  rounded corners,align=left,
  minimum width=2.5em,
  inner xsep=4pt,
  inner ysep=1pt,
  s sep= 4.3pt,
  },
  where level=1{font=\small, text width=20em,fill=blue!10}{},
  where level=2{font=\footnotesize,fill=pink!30}{},
  where level=3{font=\footnotesize,yshift=0pt,fill=yellow!20}{},
  [Flow Matching \\ Meets Biology \\ \& Life Science,fill=green!20
        [Flow Matching \\ Basics,text width=5.8em
            [General FM\\
            [\emph{Conditional FM}: \cite{albergo2023building} /
             \cite{eijkelboom2024variational} /
             \cite{lipman2022flow} /
            \cite{tong2024improving}]
            [\emph{Rectified FM}: 
            % \cite{liu2022flow} /
            \cite{lee2024improving} /
            \cite{liu2023flow} / 
            \cite{tong2024improving} /
            \cite{kornilov2024optimal}]
            [\emph{Non-Euclidean FM}: 
            \cite{chen2024flow} /
            \cite{cheng2025alpha} /
            \cite{davis2024fisher} /
            \cite{lou2020neural} /
            \cite{mathieu2020riemannian}
            ]
            ]
            [Discrete FM
            [\emph{CTMC}: \cite{campbell2024generative} / 
            \cite{cheng2025alpha} / 
            \cite{davis2024fisher} /
            \cite{gat2024discrete} / 
            \cite{qin2025defog} / 
            \cite{shaul2025flow}
            ]
            [\emph{Simplex}: 
            \cite{dunn2024mixed} /
            \cite{stark2024dirichlet} /
            \cite{tang2025gumbel}
            ]
            ]
        ]
        [Bio Sequence \\Modeling,text width=7.5em
          [DNA Sequence
            [\cite{davis2024fisher} / 
            \cite{stark2024dirichlet} /
            \cite{tang2025gumbel}
            ]
            ]
          [RNA Sequence
            [\cite{gao2024rnacg} /
            \cite{nori2024rnaflow} /
            \cite{rubin2025ribogen} / \cite{mariboflow}
            ]
            ]
          [Whole-Genome
            [\cite{klein2024genot} /
            \cite{palma2024cellflow} /
            \cite{palmamulti}]
            ]
          [Antibody Sequence
           [
           \cite{nagaraj2024igflow} / 
           \cite{tan2025dyab}
           ] 
            ]
        ]
        [Molecule Generation\\ and Design ,text width=7.5em
          [2D
              [
              \cite{DBLP:journals/corr/abs-2411-05676} / \cite{DBLP:conf/nips/EijkelboomBNWM24} /
              \cite{qin2025defog}
              ]
          ]
            [3D Modeling
                [\emph{SE(3)-equivariant:} 
                    \cite{DBLP:journals/corr/abs-2412-11082} / \cite{DBLP:conf/nips/SongGXCLEZM23} / \cite{reidenbach2024applications} / \cite{equivariantvariationalfm}
                ]
                [\emph{Efficiency:}
                    \cite{hongaccelerating} / \cite{DBLP:journals/corr/abs-2406-07266} / \cite{irwin2025semlaflow} / 
                    \cite{caoefficient} /
                    \cite{DBLP:conf/nips/HassanSLSTB24}
                ]
                [\emph{Guided Generation:}
                    \cite{DBLP:conf/nips/JiaoK0024} / \cite{DBLP:journals/corr/abs-2402-18839} / \cite{DBLP:journals/corr/abs-2404-19739} / 
                    \cite{zhouenergy2025} /
                    \cite{DBLP:journals/corr/abs-2410-18070}
                ]
            ]
          [Conditional Molecule \\ Design and Applications
              [
               \cite{DBLP:conf/nips/ZhangWL24} / 
               % \cite{DBLP:journals/jcisd/JonesKF25} / 
               \cite{DBLP:journals/corr/abs-2410-03655} / \cite{vost2024improving} / \cite{DBLP:journals/corr/abs-2412-12540} /
               \cite{zeng2025propmolflow} / \cite{bergues2025template} / \cite{zhou2025prior}
              ]
          ]
        ]
        [Protein Generation,text width=7.5em
          [Unconditional Generation
              [\emph{Backbone Generation}%: \cite{yim2023fast} / \cite{ahern2025atom} / \cite{bose2023se} / \cite{wagner2024generating} / \cite{huguet2024sequence} / \cite{geffner2025proteina} / \cite{campbell2024generative}
              ]
              [\emph{Co-design Generation}%: \cite{yang2025co} / \cite{campbell2024generative} / \cite{chen2025an}
              ]
          ]
          [Conditional Generation,
            [
            \emph{Motif-scaffolding Generation}%: \cite{yim2024improved} / \cite{yim2023fast} / \cite{ahern2025atom} / \cite{huangeva}
            ]
            [
            \emph{Pocket \& Binder Design}%: \cite{liu2024design} / \cite{cremer2025flowr} / \cite{cremer2025flowrroot} / \cite{stark2024harmonic} / \cite{zhang2024generalized} /\cite{DBLP:journals/jcisd/JonesKF25}
            ]
          ]
          [Structure Prediction
            [\emph{Conformer Prediction}
            % : \cite{jing2024alphafold} / \cite{jin2025p2dflow}
            ]
            [\emph{Side-chain Packing}
            ]
            [\emph{Docking Prediction}
            ]
          ]
          [Peptide and Antibody Generation
          ]
        ]
        [Other Biology \\ Applications ,text width=8.5em
            [Dynamic Cell Trajectory, text width=8em
              [
              \cite{zhang2025cellflux} /
              \cite{klein2024genot} / 
              \cite{kapusniak2024metric} /
              \cite{diversifiedfm}
              ]
            ]
            [Bio-Image Generation \\ and Enhancement, text width=9em
                [\cite{DBLP:journals/corr/abs-2405-18087} / \cite{DBLP:journals/corr/abs-2503-00266} / 
                \cite{DBLP:journals/cbm/ZhangHXD24}
                ]
            ]
            [Cellular Microenvironments from \\Spatial Transcriptomics
                [
                    \cite{huang2025scalable} / \cite{haviv2025wasserstein}
                ]
            ]
            [Neural Activities
                [
                    \cite{wei2025streamlevel} / \cite{wang2025flow} / \cite{collas2025riemannian}
                ]
            ]
            ]
        ]
    ]
\end{forest} 
	}
	\caption{Overview of the survey taxonomy. We begin by introducing the foundations of flow matching, including its core models and variants. Our taxonomy then categorizes flow matching applications into major biological domains and tasks. 
    }
	\label{fig: main}
\end{figure*}
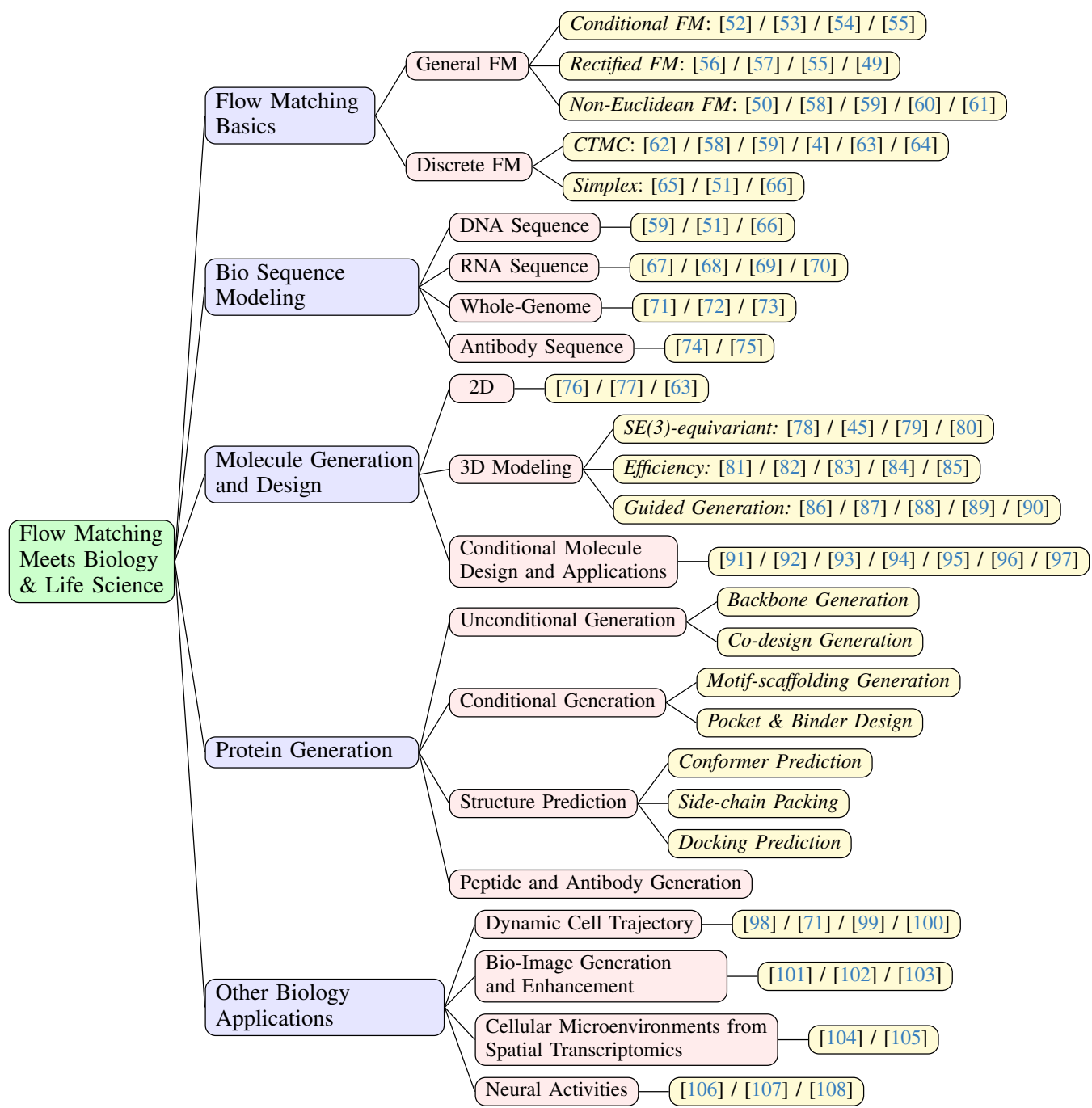

\subsection*{Challenges of Generative Modeling for Biology}
\label{sec: intro: challenges}
Biological systems are among the most intricate and multifaceted systems in the natural world~\cite{ruth1997modeling, edelman2001degeneracy, haefner2005modeling}, shaped by billions of years of evolution and governed by deeply intertwined physical, chemical, and informational processes. Modeling such systems has long been a grand challenge across scientific disciplines, demanding tools that can reconcile precision with flexibility \cite{rhie2021towards, one2019one, kim2020architecture, sahin2014mrna, baker2001protein, baek2021accurate}. The complexity of biological data and phenomena stems from a confluence of factors, with some of the most formidable challenges including: (i) the necessity to embed \underline{rich domain knowledge}, ranging from physical laws to biochemical constraints, into generative models in a way that ensures structural and functional validity; (ii) the \underline{scarcity, incompleteness, and noise} characteristic of real-world biological \underline{datasets}, often resulting from expensive or error-prone experimental procedures; (iii) the inherently \underline{multi-scale and multi-modal} nature of biological processes, which span atomic interactions to cellular behavior, and integrate diverse data types such as sequences, structures, and spatial-temporal signals; (iv) the increasing demand for \underline{controllable and condition-aware} generation, where outputs must satisfy explicit biological properties or therapeutic objectives; and (v) the pressing need for models that are not only accurate but also computationally \underline{scalable and sample-efficient}, especially in applications such as drug discovery or protein design where inference speed can be critical. Together, these challenges make it challenging for biology models.

FM, as a recently introduced generative modeling paradigm, holds strong potential for addressing the unique challenges of biological data. It learns a deterministic vector field to map a simple base distribution directly to complex target data via continuous probability trajectories. This yields several advantages particularly relevant to biological applications, such as faster and more stable sampling, easier conditioning on structured inputs, and the ability to incorporate geometric or physical priors into the modeling process. 
Since its introduction, a growing number of studies have explored the use of FM in tackling biological tasks. 
These early successes demonstrate not only the method’s versatility but also its capacity to model the structured, multimodal, and constraint-rich nature of biological systems, positioning FM as a compelling alternative to conventional generative frameworks in the life sciences.

\subsection*{Our Contributions}
This survey presents the first comprehensive review of FM and its applications in  biology and life sciences. Our key contributions are summarized as follows:

\begin{itemize}
    \item \textbf{A Unified Taxonomy of Flow Matching Variants.} We introduce a structured taxonomy of FM methodologies, spanning general FM, conditional and rectified FM, non-Euclidean and discrete FM, and hybrid variants.

    \item \textbf{In-depth Survey of Biological Applications.} We systematically categorize and review the use of FM across three primary biological domains: \emph{biological sequence modeling}, \emph{molecule generation and design}, and \emph{protein generation}. We further explore several other emerging applications beyond this scope.

    \item \textbf{Comprehensive Benchmark and Dataset Survey.} We compile and review widely used biological datasets, benchmarks, and software tools adopted in FM research.

    \item \textbf{Trend, Challenges, and Emerging Directions.} We contextualize the evolution of FM through bibliometric trends and identify key methodological innovations. We further analyze domain-specific modeling challenges which may motivate new FM research directions.

    \item \textbf{Bridging Modeling and Biology Communities.} By mapping methodological advances in FM to diverse biological challenges, we offer a cross-disciplinary bridge that connects the machine learning community developing FM algorithms with the biological sciences community seeking powerful generative tools.

\end{itemize}

\subsection*{Connection to Existing Survey}
\label{sec: intro: connection to existing}

\begin{table*}[t!]
\centering
\caption{Existing surveys related to this work. We present the first comprehensive survey \\ dedicated to flow matching and its applications in biology and life sciences.}
\label{tab: comparison of existing surveys}
\begin{tabular}{lll}
\toprule
Reference &
  Generative Modeling &
  Task Domain \\ \midrule
Jabbar et al. (2021) \cite{jabbar2021survey} &
  Generative Adversarial Network &
  General \\
Xia et al. (2022) \cite{xia2022gan} &
  Generative Adversarial Network &
  Anomaly Detection\\
Greener et al. (2022) \cite{greener2022guide} &
  Various ML and Generative Modeling Methods &
  Biology, including protein design and DNA sequence   \\
Li et al. (2023) \cite{li2023comprehensive} &
  Autoencoder &
  General, including Image Classification and NLP Tasks\\ 
Yang et al. (2023) \cite{yang2023diffusion} &
  Diffusion Model &
  General, including CV, NLP, Multimodal Tasks\\ 
Croitoru et al. (2023) \cite{croitoru2023diffusion} &
  Diffusion Model &
  Various Tasks in Computer Vision\\ 
Liang et al. (2024) \cite{liang2024survey} &
  Variational Autoencoder &
   Recommendation \\ 
Cao et al. (2024) \cite{cao2024survey} &
   Diffusion Model &
   General, including image, video and audio Generation \\ 
Guo et al. (2024) \cite{guo2024diffusion} &
   Diffusion Model &
   Biology, including protein, molecular, gene-expression tasks \\ 
Saad et al. (2024) \cite{saad2024survey} &
  Generative Adversarial Network &
  Biomedical Image Synthesis \\ 
Tang et al. (2024) \cite{tang2024survey} &
  Various Generative Modeling Methods &
  De Novo Drug Design \\ 
Du et al. (2024) \cite{du2024machine} &
  Various Generative Modeling Methods &
  Molecular Design\\
Zhang et al. (2025) \cite{zhang2025scientific} &
  Large Language Models &
  Biology and Chemistry, e.g., molecular, protein, genomic tasks\\
Morehead et al. (2025) \cite{morehead2025go} & Flow Matching & Biology, including molecule, single \& multi-cellular, and bioimaging tasks.\\
\textbf{Ours} &
  Flow Matching &
  Various Tasks in Biology and Life Science\\ \bottomrule
\end{tabular}%
% }
\end{table*}

Existing related surveys can be broadly categorized into three groups. The first category focuses exclusively on generative modeling methodologies. These surveys either provide comprehensive overviews of specific classes of generative models \cite{li2023comprehensive, jabbar2021survey, cao2024survey} or examine their applications within particular domains, such as computer vision \cite{croitoru2023diffusion}, recommendation systems \cite{liang2024survey}, and anomaly detection \cite{xia2022gan}. The second category surveys the use of generative models in biology prior to the advent of FM. For example, \cite{du2024machine} reviews generative models for molecular design, \cite{tang2024survey} focuses on de novo drug design, and \cite{greener2022guide} provides a broad overview of machine learning methods in both predictive and generative biological modeling.
A concurrent survey~\cite{morehead2025go} emphasizes practical guidance and open-source tooling, our survey offers a unified taxonomy of flow-matching methodologies with fine-grained links to specific biological problem classes.
Table~\ref{tab: comparison of existing surveys} presents a comparison of existing surveys on generative modeling, highlighting their covered model classes and application domains.
To the best of our knowledge, this work presents the first comprehensive survey dedicated to FM and its applications in biology and life sciences. By bridging recent developments in generative modeling with their emerging applications in biological domains, this survey aims to fill a critical gap in the literature.

\subsection*{Outline of the Survey}
To provide a comprehensive understanding of FM in the context of biology and life sciences, this survey is organized into several key sections. We begin by introducing the fundamental concepts and methodologies underlying FM in Section "Flow-Matching Basics", establishing a foundation for its application in biological contexts. Next, in Section "Sequence Modeling", we delve into specific areas of application, starting with biology sequence generation, followed by molecule generation and design in Section "Molecule Generation", and then peptide and protein generation in Section "Protein Generation", each highlighting recent advancements and representative studies. In Section "Other Bio Applications", we also discuss other emerging applications of FM in biology. Finally, we conclude by outlining future research directions and potential challenges, aiming to inspire further exploration and innovation in this rapidly evolving field. Figure \ref{fig: main} presents the overall structure of this survey, with each section divided into various subtopics for a more detailed exploration.

\section*{Background}
Generative modeling seeks to learn a probability distribution \( p_{\text{data}}(x) \) from a dataset of examples \( \{x_i\}_{i=1}^N \), such that we can generate new samples \( \hat{x} \sim p_{\theta}(x) \) that resemble real data. 
These models underpin advances in biology tasks ranging from molecular generation to protein design and cellular imaging \cite{tang2024survey, du2024machine, mock2024recent, kell2020deep, liu2024geometric}, with AlphaFold \cite{abramson2024accurate, yang2023alphafold2, jumper2021highly} standing out as one of the most prominent and transformative examples, recognized with the Nobel Prize in 2024. 
AlphaFold leverages deep generative principles to predict protein 3D structures directly from amino acid sequences, a task that had challenged the field for decades \cite{mariani2013lddt, baek2021accurate, baek2023efficient}. 
By effectively modeling the conditional distribution over protein conformations, AlphaFold not only revolutionized protein structure prediction but also highlighted the broader potential of generative models to capture complex, structured biological phenomena at scale. 
In biology domains, data is often high-dimensional, multimodal, and governed by physical or biochemical constraints \cite{berman2003announcing, shimoyama2015rat, alquraishi2019proteinnet, kim2016pubchem}, requiring generative models to strike a careful balance between validity, diversity, and interpretability. 
In this section, we provide a brief overview of the major paradigms in generative modeling, with the goal of establishing a conceptual and mathematical foundation for understanding more recent developments such as FM.
For clarity and consistency, all symbols used throughout this paper are summarized in Table~\ref{tab:notation}. We also briefly compare different generative modeling paradigms and FM in Table \ref{tab: comparison of generative modeling}. To further enhance accessibility for readers from diverse scientific backgrounds, we provide a glossary of key technical terms in the Supplementary Information Section "Technical Terms".

\begin{table}
\centering
\caption{Notation used in generative modeling paradigms. A glossary of technical terms is provided in the Supplementary Information Section "Technical Terms".}
\label{tab:notation}
\begin{tabular}{@{}cl@{}}
\toprule
\textbf{Symbol} & \textbf{Description} \\ \midrule
\( x \) & Data sample \\
\( z \) & Latent variable \\
\( p_{\text{data}}(x) \) & True data distribution \\
\( p_\theta(x) \) & Model distribution \\
\( f_\theta \) & Invertible function (flow) \\
\( u_\theta(x, t) \) & Velocity field in FM \\
\( p_t(x) \) & Intermediate distribution at time \( t \) \\
\( \epsilon \) & Noise in diffusion model \\
\( \mathcal{L}_{\text{FM}} \) & Flow Matching loss \\
\( \mathcal{L}_{\text{DM}} \) & Diffusion model loss \\
\bottomrule
\end{tabular}
\end{table}
\subsection*{Variational Autoencoder (VAE)}

Variational Autoencoders (VAEs) \cite{kingma2013auto, diederik2019introduction, girin2022dynamical, pu2016variational, kusner2017grammar} are a class of latent-variable generative models that aim to model the data distribution \( p_{\text{data}}(x) \) through a learned probabilistic decoder \( p_\theta(x|z) \), where \( z \) is a latent variable drawn from a prior \( p(z) \), typically a standard Gaussian. Since the true posterior \( p(z|x) \) is often intractable, VAEs introduce an approximate posterior \( q_\phi(z|x) \), known as the encoder, and optimize the model using variational inference.
The training objective is to maximize a variational lower bound, known as the evidence lower bound (ELBO), on the marginal log-likelihood of the data:
\begin{equation}
\log p_\theta(x) \geq \mathbb{E}_{q_\phi(z|x)} \left[ \log p_\theta(x|z) \right] - \mathrm{KL}\left( q_\phi(z|x)\| p(z) \right)
\end{equation}
The first term encourages accurate reconstruction of the input data from the latent variable \( z \), while the second term regularizes the approximate posterior to stay close to the prior distribution. During training, the reparameterization trick is used to allow gradients to backpropagate through the sampling process, typically by expressing \( z \sim q_\phi(z|x) \) as \( z = \mu(x) + \sigma(x) \odot \epsilon \), where \( \epsilon \sim \mathcal{N}(0, I) \).
However, VAEs often suffer from over-regularization and produce blurred outputs, especially in high-dimensional domains such as images and molecular graphs \cite{bredell2023explicitly, takida2022preventing, dai2020usual}. 

\subsection*{Generative Adversarial Network (GAN)}

Generative Adversarial Networks (GANs) \cite{goodfellow2020generative} are a class of implicit generative models that learn to generate realistic data by playing a two-player minimax game between two neural networks: a generator \( G_\theta \) and a discriminator \( D_\phi \). The generator maps noise samples \( z \sim p(z) \), typically drawn from a simple prior such as a Gaussian, into synthetic data samples \( G_\theta(z) \). The discriminator attempts to distinguish between real samples \( x \sim p_{\text{data}} \) and generated samples \( G_\theta(z) \).
The original GAN objective is formulated as:
\begin{equation}
\min_{G_\theta} \max_{D_\phi} \; \mathbb{E}_{x \sim p_{\text{data}}} [\log D_\phi(x)] + \mathbb{E}_{z \sim p(z)} [\log (1 - D_\phi(G_\theta(z)))]
\end{equation}

GANs are known to suffer from several practical challenges, including training instability, sensitivity to hyperparameters, and mode collapse 
Numerous variants have been proposed to improve training dynamics and sample diversity, such as Wasserstein GANs \cite{DBLP:journals/corr/ArjovskyCB17}, Least-Squares GANs \cite{mao2017least}, and conditional GANs \cite{mirza2014conditional}. In biological applications, GANs have been used for generating realistic cell images \cite{bafti2023biogan}, synthesizing gene expression profiles \cite{chaudhari2020data, lee2023recent}, and augmenting scarce datasets \cite{yang2025improved}. Despite their limitations, their ability to capture complex data distributions without explicit density estimation makes them a compelling choice for modeling high-dimensional biological data \cite{osokin2017gans}.

\subsection*{Flow-Based Model}
\label{sec: normalizing flow}
Flow-based models (also known as normalizing flows) \cite{rezende2015variational, kobyzev2020normalizing} are a family of generative models that construct complex data distributions by applying a sequence of invertible transformations to a simple base distribution, typically a standard Gaussian distribution. Given a base variable \( z \sim p_z(z) \), a flow model learns an invertible mapping \( x = f_\theta(z) \) such that the model distribution \( p_\theta(x) \) can be computed exactly via the change-of-variables formula:
\begin{equation}
\log p_\theta(x) = \log p_z(f_\theta^{-1}(x)) + \log \left| \det \left( \frac{\partial f_\theta^{-1}(x)}{\partial x} \right) \right|
\end{equation}

The goal is to train the parameters \( \theta \) to maximize the log-likelihood of the observed data under this model. The invertibility of \( f_\theta \) allows for exact and tractable likelihood computation, efficient sampling, and deterministic inference.
To ensure both tractability and expressivity, flow models are often constructed as a composition of multiple simple bijective transformations:
\begin{equation}
f_\theta = f_K \circ f_{K-1} \circ \cdots \circ f_1\label{eq:stacked normalizing flow}
\end{equation}
Each component \( f_k \) is designed to allow efficient computation of the Jacobian determinant and its inverse. Representative architectures include NICE~\cite{DBLP:journals/corr/DinhKB14},  RealNVP~\cite{DBLP:conf/iclr/DinhSB17}, Glow~\cite{DBLP:conf/nips/KingmaD18}, and Masked Autoregressive Flows (MAF)~\cite{DBLP:conf/nips/PapamakariosMP17}, which utilize affine coupling layers or autoregressive transforms to maintain invertibility.

However, the invertible constraint on $f_\theta$ along with the need to compute the determinant of the Jacobian $\frac{\partial f_\theta(x)}{\partial x}$ imposes significant constraints on model expressiveness and design flexibility. Continuous normalizing flow (CNF)~\cite{chen2018neural} address these limitations by replacing the discrete sequence of transformations (Eq.~\eqref{eq:stacked normalizing flow}) with a continuous-time dynamic system $\frac{dx}{dt} = f(x(t),t)$. This formulation leads to a more efficient computation of the log-density change:
\begin{equation}
    \frac{\partial \log p(x(t))}{\partial t} = -\text{Tr}\left(\frac{df}{dx(t)}\right)\label{eq:CNF log p ode}
\end{equation}
Notably, the vector field $f$ is not required to be invertible.

CNFs serve as a foundational building block for FM. While CNFs allow for more expressive modeling, their training via maximum likelihood still demands computationally expensive ODE solvers. A core motivation behind flow matching is to simplify the training of ODE-based generative models, without sacrificing the benefits of continuous-time formulations.

\subsection*{Diffusion Models (DM)}

Diffusion models~\cite{DBLP:conf/icml/Sohl-DicksteinW15,ho2020denoising,DBLP:conf/nips/SongE19,DBLP:conf/iclr/0011SKKEP21,DBLP:conf/iclr/SongME21} are a family of likelihood-based generative models that generate data by reversing a gradual noising process. They define a forward process that incrementally transforms data into noise, and parameterize a neural network to fit the groundtruth reverse process, recovering data from noise step by step.

\paragraph{Forward Process} The forward process defines a sequence of latent variables \( \{x_t\}_{t=0}^T \), which are the gradually corrupted version of the clean data \( x_0 \sim p_{\text{data}} \). A typical forward process is formulated as a set of Gaussian distributions conditioned on the previous step: 
\begin{equation}
q(x_t | x_{t-1}) = \mathcal{N}(x_t; \sqrt{1 - \beta_t} x_{t-1}, \beta_t I)
\end{equation}
where $\{\beta_t\}$ is called noise schedule. Usually, the distribution of the corrupted data at any time $t$ has a closed form:
\begin{align}
q(x_t | x_0) &= \mathcal{N}(x_t; \sqrt{\bar{\alpha}_t} x_0, (1 - \bar{\alpha}_t) I),\\ \bar{\alpha}_t &= \prod_{s=1}^t (1 - \beta_s)
\end{align}

\paragraph{Training} Similar to many likelihood-based models, negative log-likelihood is a canonical choice of the loss function~\cite{DBLP:conf/icml/Sohl-DicksteinW15,ho2020denoising,DBLP:conf/nips/CampbellBBRDD22}. Beyond that, cross-entropy or square error are also widely used~\cite{DBLP:conf/nips/AustinJHTB21,ho2020denoising}. Based on that, neural networks (NNs) are used to parameterize various components of the diffusion process, such as to predict the data~\cite{DBLP:conf/iclr/SalimansH22}, predict the noise~\cite{ho2020denoising}, and predict the score~\cite{DBLP:conf/iclr/SongME21}. The following unweighted regression loss for predicting the noise is a popular example:
\begin{align}
\mathcal{L}_{\text{DM}} &= \mathbb{E}_{x_0, t, \epsilon} \left[ \left\| \epsilon - \epsilon_\theta(x_t, t) \right\|^2 \right]\\
x_t &= \sqrt{\bar{\alpha}_t} x_0 + \sqrt{1 - \bar{\alpha}_t} \epsilon,\ \epsilon \sim \mathcal{N}(0, I)
\end{align}
% where \( \epsilon \sim \mathcal{N}(0, I) \) is standard Gaussian noise.

\paragraph{Generation} Equipped with the NN-parameterized component, the reverse process of the diffusion process is used for generation. For example, the reverse process with the NN-predicted noise $\epsilon_\theta$ can denoise the Gaussian noise \( x_T \sim \mathcal{N}(0, I) \) gradually:
\begin{align}
x_{t-1} &= \frac{1}{\sqrt{1 - \beta_t}} \left( x_t - \frac{\beta_t}{\sqrt{1 - \bar{\alpha}_t}} \epsilon_\theta(x_t, t) \right) + noise
\end{align}

A well-known limitation of diffusion models is their slow sampling process, which often requires hundreds of iterative steps. To address this inefficiency, several acceleration techniques have been proposed, including the adoption of tailored numerical solvers~\cite{DBLP:conf/nips/0011ZB0L022}, model distillation~\cite{DBLP:conf/iclr/SalimansH22}, and continuous-time formulations~\cite{DBLP:conf/iclr/SongME21,DBLP:conf/nips/CampbellBBRDD22}. Notably, Probability flow ODE~\cite{DBLP:conf/iclr/0011SKKEP21} and DDIM~\cite{DBLP:conf/iclr/SongME21} demonstrate that there exists a deterministic ODE whose solution shares the same marginal distributions as the reverse-time stochastic differential equation (SDE) used in diffusion models. This observation is conceptually aligned with the idea behind flow matching (FM), although both probability flow ODE and DDIM remain trained using the standard loss functions of diffusion models, such as the evidence lower bound (ELBO).

\subsection*{Consistency Models}

\setcounter{paragraph}{0}

Consistency models (CMs)~\cite{song2023consistency} are a recent family of generative models built upon the diffusion models. They aim to bypass the slow iterative denoising procedure of diffusion sampling by learning a direct mapping from noise to data. 

\paragraph{Forward Process}
A consistency model is a neural function $f_{\theta}(x_t, t)$ that approximates the solution of the Probability flow ODE (PF-ODE) in closed form. Given a noisy sample $x_t$ at time $t$, $f_\theta$ predicts its corresponding clean data $x_0$. A defining property of CMs is \emph{self-consistency}: all points on the same diffusion trajectory should map to the same output. 

\paragraph{Training}
CMs are trained from two main paradigms: Consistency Distillation and Consistency Training.

Consistency Distillation (CD) \cite{song2023consistency} distills a pretrained diffusion teacher into $f_\theta$. Given adjacent states $(x_t, x_{t+\Delta})$ along the teacher’s PF-ODE trajectory, the student minimizes
\begin{equation}
\mathcal{L}_{\text{CD}} =
\mathbb{E}\!\left[\!\left\|
f_\theta(x_{t+\Delta}, t+\Delta)
 - f_\theta(x_t, t)
\right\|_2^2\!\right]
\end{equation}

Consistency Training (CT) \cite{song2024improved, song2023consistency} trains $f_\theta$ from scratch without a teacher by sampling two noisy versions $(x_s, x_t)$ of the same data $x_0$ via a shared noise realization $z$:
$x_t = x_0 + \sigma(t)z$, $x_s = x_0 + \sigma(s)z$: 
\begin{equation}
\mathcal{L}_{\text{CT}} =
\mathbb{E}\!\left[\!\left\|
f_\theta(x_t,t) - f_\theta(x_s,s)
\right\|_2^2\!\right]
\end{equation}

Beyond the original formulation~\cite{song2023consistency}, several variants have extended this idea. Multi-step CMs~\cite{heek2024multistep} refine generation by repeatedly evaluating $f_\theta$ over decreasing times $(t_n \!\to\! 0)$. In addition, diffusion models are integrated with consistency models ~\cite{kim2024consistency,geng2025consistency}. 
Some recent approaches further emphasize later noise stages during training \cite{lee2025truncated}.

\begin{table*}[h]
\centering
\caption{Comparison of major generative modeling paradigms.}
\label{tab: comparison of generative modeling}
\resizebox{.8\textwidth}{!}{%
\begin{tabular}{>{\centering\arraybackslash}m{3.0cm}
                >{\centering\arraybackslash}m{2.5cm}
                >{\centering\arraybackslash}m{3.8cm}
                >{\centering\arraybackslash}m{4.3cm}
                }
\toprule
\textbf{Model Type} & 
\textbf{Training Objective} & 
\textbf{Number of Function Evaluations} & 
\textbf{Structured Data Support} \\
\midrule

\textbf{VAE} & 
Likelihood & 
Low & 
Moderate (via extensions) \\

\textbf{GAN} & 
Adversarial loss & 
Low & 
Weak (limited geometry) \\

\textbf{Diffusion} & 
Likelihood & 
SDE solver-dependent & 
Strong (SE(3), graph diffusion) \\

\textbf{Consistency Model} & 
Likelihood & 
SDE solver-dependent & 
Strong (SE(3), graph diffusion) \\

\textbf{Flow-Based} & 
Likelihood & 
Low & 
Moderate (design-dependent) \\

\textbf{Flow Matching} & 
Velocity matching & 
ODE solver-dependent & 
Strong (geometry-aware, equivariant) \\

\bottomrule
\end{tabular}
}
\end{table*}
\section*{Flow-Matching Basics}\label{sec:fm}

In this section, we provide background knowledge on flow-matching (FM) models, including general FM and discrete FM.

\subsection*{General Flow-Matching}
\setcounter{paragraph}{0}
Flow-matching is a continuous-time generative framework that generalizes diffusion models by \textit{regressing a vector field that transports one distribution into another}~\cite{lipman2022flow}. In general, FM aims to construct a velocity field \( u_\theta(x, t) \) to transport a source \( p_0 \) to a target \( p_1 \) via the continuity equation:
\begin{equation}
\frac{\partial p_t}{\partial t} + \nabla \cdot (p_t u_\theta(x, t)) = 0.
\end{equation}
An FM can be trained by minimizing the squared loss between the neural velocity field $u_\theta(x,t)$ and a reference velocity field $u^*_t(x,t)$ as follows
\begin{equation}\label{eq:fm_obj}
    \mathcal{L}_{\text{FM}} = \E_{t\sim[0,1], x_t\sim p_t(x)}\|u^*(x_t,t)-u_\theta(x_t,t)\|^2.
\end{equation}

Promising as it might be, directly optimizing the objective in Eq.~\eqref{eq:fm_obj} is impractical: the optimal velocity field \(u^*(x,t)\) encodes a highly complex joint transformation between two high-dimensional distributions~\cite{lipman2024flow}. To overcome this challenge, conditional FM variants have been introduced to enable more tractable training (Paragraph~\ref{para:conditional}). Concurrently, rectified FM methods propose improved noise couplings along the straight‐line probability path (Paragraph~\ref{para:rectified}). Finally, non-Euclidean FM extensions generalize the framework from flat Euclidean space to curved manifolds, accommodating data with intrinsic geometric structure (Paragraph~\ref{para:non-euc}).

\paragraph{Conditional FM~\cite{lipman2022flow,eijkelboom2024variational,albergo2023building,tong2024improving}}\label{para:conditional}
To resolve the intractable $u^*(x,t)$, conditional FM introduces a conditioning variable $z$, e.g., class label, and define a conditional path $p(x|t,z)$ such that the induced global path $p(x|t)=\int_{z}p(x|t,z)p(z)dz$ transforms $p_0$ to $p_{\text{data}}$ and the corresponding conditional velocity field has analytical form. A conditional FM can be trained by minimizing the quadratic loss between the neural velocity field $u_\theta(x,t)$ and the conditional velocity field $u^*_t(x,t,z)$ as follows
\begin{equation}\label{eq:cfm_obj}
    \E_{t\sim[0,1], x_t\sim p_t(x\mid z), z\sim p_z}\|u^*(x_t,t,z)-u_\theta(x_t,t)\|^2.
\end{equation}
The training procedure involves sampling a conditioning variable $z$, e.g., via linear interpolation~\cite{albergo2023building,liu2023flow} or Gaussian path~\cite{lipman2022flow}, and random time $t$, constructing $x_t$ along the prescribed path, and minimizing the corresponding loss.
Once the model is trained, the sampling/generation process is done by solving the learned ODE $dx/dt = u_\theta(x,t)$ using an ODE solver from $t=0$ (noise) to $t=1$ (data).
The key theoretical foundation of conditional FM is that the gradient of the FM objective in Eq.~\eqref{eq:fm_obj} is equivalent to gradient of the CFM objective in Eq.~\eqref{eq:cfm_obj}.
Building upon the conditioning variable $z$, one can define velocity field in analytical forms with tractable training.

\paragraph{Rectified FM~\cite{liu2023flow,lee2024improving,tong2024improving,kornilov2024optimal}}\label{para:rectified}
Infinite probability path exist between source and target distributions that can be leveraged by conditional FM, rectified FM prefers the linear transport trajectory that best connect two distributions. \cite{liu2023flow} proposes to train a velocity field carrying each sample $x_0$ to its paired target $x_1$ along nearly-straight lines via
\begin{equation}
    \E_{(x_0,x_1)\sim \pi}\int_0^1\|u_\theta(x_t,t)-(x_1-x_0)\|^2dt
\end{equation}
where $pi$ is a coupling of $p_0$ and $p_1$. It is shown that the optimal transport (OT) coupling provides a straight coupling for $p_0$ and $p_1$, simplifying the flow and reducing curliness~\cite{liu2023flow,tong2024improving}.

\paragraph{Non-Euclidean FM~\cite{chen2024flow,davis2024fisher,lou2020neural,mathieu2020riemannian}}\label{para:non-euc}
Non-Euclidean flows extend continuous flows to curved data spaces. For example, \cite{mathieu2020riemannian} introduce Riemannian Continuous Normalizing Flows, defining the generative flow by an ODE on the manifold to model flexible densities on spheres, tori, hyperbolic spaces, etc..
\cite{lou2020neural} propose Neural Manifold ODEs, integrating dynamics chart-by-chart (e.g. via local coordinate charts) so that the learned velocity field stays tangent to the manifold.
More recently, \cite{chen2024flow} propose Riemannian FM by using geodesic distances as a “premetric” they derive a closed-form target vector field pushing a base distribution to the data without any stochastic diffusion or divergence term.
On simple manifolds (e.g. spheres or hyperbolic space where geodesics are known) Riemannian FM is completely simulation-free, and even on general geometries it only requires solving a single ODE without calculating expensive score or density estimates.
\cite{davis2024fisher} introduce Fisher FM, treating categorical distributions as points on the probability simplex with the Fisher–Rao metric and transporting them along spherical geodesics.
In general, Riemannian flows replace straight-line interpolations with intrinsic geodesics and explicitly account for the manifold’s metric (e.g. via the Riemannian divergence in the change-of-density). These works tackle the challenges of defining tangent vector fields and volume corrections on curved spaces via chart-based integration, metric-adjusted log-density formulas, or flow-matching losses that avoid divergence estimates. Overall, they enable scalable generative modeling on curved domains (spheres, Lie groups, statistical manifolds, etc.), respecting curvature in ways standard Euclidean FM cannot.

\subsection*{Discrete Flow-Matching}
\setcounter{paragraph}{0}
Discrete FM has emerged as a powerful paradigm for generative modeling over discrete data domains, such as sequences, graphs, and categorical structures, covering a wide range of biological objects~\cite{DBLP:conf/nips/AustinJHTB21,gat2024discrete}. By extending the principles of continuous FM to discrete spaces, DFM enables the design of efficient, non-autoregressive generative models. This section delves into two principal frameworks: Continuous-Time Markov Chain (CTMC)-based methods (Paragraph~\ref{para:ctmc}) and simplex-based methods (Paragraph~\ref{para:simplex}).

\paragraph{Continuous-Time Markov Chain (CTMC)}\label{para:ctmc}
CTMC-based approaches model the generative process as a continuous-time stochastic evolution over discrete states, leveraging the mathematical framework of continuous-time Markov chains to define and learn probability flows.
\cite{campbell2024generative} utilizes CTMCs to model flows over discrete state spaces. This approach allows for the integration of discrete and continuous data, facilitating applications like protein co-design by enabling multimodal generative modeling.
Fisher Flow~\cite{davis2024fisher} adopts a geometric perspective by considering categorical distributions as points on a statistical manifold endowed with the Fisher-Rao metric. This approach leads to optimal gradient flows that minimize the forward Kullback-Leibler divergence, improving the quality of generated discrete data.
\cite{shaul2025flow} expanded the design space of discrete generative models by allowing arbitrary discrete probability paths within the CTMC framework. This holistic approach enables the use of diverse corruption processes, providing greater flexibility in modeling complex discrete data distributions.
DeFog~\cite{qin2025defog} is a discrete FM framework tailored for graph generation. By employing a CTMC-based approach, DeFoG achieves efficient training and sampling, outperforming existing diffusion models in generating realistic graph.

\paragraph{Simplex-based discrete FM}\label{para:simplex}
Simplex-based methods operate within the probability simplex, modeling flows over continuous relaxations of discrete distributions. These approaches often employ differentiable approximations to handle the challenges posed by discrete data.
SimplexFlow~\cite{dunn2024exploring} combines continuous and categorical flow matching for 3D de novo molecule generation, where intermediate states are guaranteed to reside on the simplex.
Dirichlet FM~\cite{stark2024dirichlet} utilizes mixtures of Dirichlet distributions to define probability paths over the simplex, addressing discontinuities in training targets and enables efficient.
$\alpha$-flow~\cite{cheng2025alpha} unifies various continuous-state discrete FM models under the lens of information geometry. By operating on different $\alpha$-representations of probabilities, this framework optimizes the generalized kinetic energy, enhancing performance in tasks such as image and protein sequence generation.
STGFlow~\cite{tang2025gumbel} employs a Gumbel-Softmax interpolant with a time-dependent temperature for controllable biological sequence generation, which includes a classifier-based guidance mechanism that enhances the quality and controllability of generated sequences.
\section*{Sequence Modeling}\label{sec:sequence}
FM has emerged as a powerful framework for biological sequence generation, offering deterministic and controllable modeling of discrete structures such as DNA, RNA, and whole-genome data. In this section, we survey different FM models designed for biological sequence generation, including DNA sequence, RNA sequence, whole-genome modeling, and antibody design. 
By leveraging continuous transformations, flow matching enables efficient generation of sequences conditioned on various biological constraints and properties.

\subsection*{DNA Sequence Generation}
Early deep generative models, e.g. GANs or autoregressive models, struggled to satisfy the complex constraints of functional genomics sequences. 
FM models provide natural solutions to bridge this gap by mapping discrete nucleotide sequences into continuous probabilistic spaces for training~\cite{stark2024dirichlet}.
Instead of simulating a stochastic diffusion~\cite{stark2024dirichlet}, FM models directly train a continuous vector field that transports a simple base distribution, e.g., uniform distribution over nucleotides, into the empirical DNA data distribution.

Fisher-Flow~\cite{davis2024fisher} introduces a geometry-based flow matching approach, which treats discrete DNA sequences as points on a statistical manifold endowed with the Fisher–Rao metric.
By allowing for continuous reparameterization of discrete data, probability mass is transported along optimal geometric paths on the positive orthant of a hypersphere, achieving state-of-the-art performance on DNA promoter and enhancer sequence generation benchmarks compared to earlier diffusion-based and flow-based models.

Besides categorical distribution, Dirichlet distribution is adopted to handle discrete sequences. Dirichlet Flow~\cite{stark2024dirichlet} utilizes mixtures of Dirichlet distributions to define probability paths on the simplex, addressing discontinuities and pathologies in naive linear flow matching. Dirichlet Flow enables one-step DNA sequence generation and achieves superior distributional metrics and target-specific design performance compared to prior models on complex DNA design tasks.

In addition, STGFlow~\cite{tang2025gumbel} proposes straight-through guidance, combining Gumbel-Softmax flows with classifier-based guidance to steer the generation process toward desired sequence properties, facilitating controllable de novo DNA sequence generation. MOG-DFM \cite{DBLP:journals/corr/abs-2505-07086} generalizes discrete flow matching guidance into a multi-objective paradigm. It leverages multiple scalar objectives and computes a hybrid rank-directional score at each sampling step.

\subsection*{RNA Sequence Generation}
Flow matching has recently been applied to RNA sequence and structure design.
Rather than focusing solely on sequence generation, existing FM methods prioritize structural fidelity, enabling advanced applications in inverse folding, protein-conditioned design, and ensemble backbone sampling.
RNACG~\cite{gao2024rnacg} introduces a versatile flow-matching framework for conditional RNA generation that supports tasks ranging from 3D inverse folding to translation efficiency prediction. 
RNAFlow~\cite{nori2024rnaflow} couples an RNA inverse-folding module with a pretrained structure predictor to co-generate RNA sequences and their folded structures in the context of bound proteins.
RiboGen~\cite{rubin2025ribogen} develops the first deep network to jointly synthesize RNA sequences and all-atom 3D conformations via equivariant multi-flow architectures. 
RNAbpFlow~\cite{tarafder2025rnabpflow} presents a SE(3)-equivariant flow-matching model that conditions on both sequence and base-pair information to sample diverse RNA backbone ensembles.
More recently, RiboFlow~\cite{mariboflow} proposes to synergize the design of RNA structure and sequence by integrating RNA backbone frames, torsion angles and sequence features for an explicit modeling on RNA's dynamic conformation.

\subsection*{Whole-Genome Modeling}
At the whole-genome level, flow matching has been applied to model single-cell genomics data. GENOT~\cite{klein2024genot} employs entropic Gromov-Wasserstein flow matching to learn mappings between cellular states in single-cell transcriptomics, facilitating studies of cell development and drug response. cellFlow~\cite{palma2024cellflow} proposes a generative flow-based model for single-cell count data that operates directly in raw transcription count space, preserving the discrete nature of the data. CFGen~\cite{palmamulti} introduces a flow-based conditional generative model capable of generating multi-modal and multi-attribute single-cell data, addressing tasks such as rare cell type augmentation and batch correction.

\subsection*{Antibody Sequence Generation}
FM has also been utilized for antibody sequence generation.
IgFlow~\cite{nagaraj2024igflow} proposes a SE(3)-equivariant FM model for de novo antibody variable region generation (heavy/light chains and CDR loops). IgFlow supports unconditional antibody sequence-structure generation and conditional CDR loop inpainting, producing structures comparable to those from a diffusion-based model while achieving higher self-consistency in conditional designs; it also offers efficiency benefits like faster inference and better sample efficiency than the diffusion counterpart. dyAb~\cite{tan2025dyab} proposes a flexible antibody design FM, which integrates coarse-grained antigen–antibody interface alignment with fine-grained flow matching on both sequences and structures. By explicitly modeling antigen conformational changes (via AlphaFold2 predictions) before binding, dyAb significantly improves the design of high-affinity antibodies in cases where target antigens undergo dynamic structural shifts.

These advancements demonstrate the versatility of flow matching in modeling complex biological sequences and structures, providing a unified framework for deterministic and controllable generation across various biological domains.

\section*{Molecule Generation}\label{sec:molecule}

Molecule generation is a fundamental task in biological modeling, playing a crucial role in drug discovery, material design, and understanding molecular interactions \cite{hoogeboom2022equivariant, DBLP:conf/nips/LiuABG18, DBLP:conf/iclr/VignacKSWCF23}. The ability to generate novel molecules with desired properties has significant implications for both theoretical and applied research in life sciences \cite{DBLP:conf/nips/LuoGMP21, peng2022pocket2mol}. Traditional approaches, such as rule-based simulations and heuristic algorithms, often face challenges in scalability and diversity \cite{noe2020machine, hollingsworth2018molecular}. In contrast, generative models, including flow matching, offer a data-driven approach to efficiently explore the vast chemical space \cite{walters2020applications, guo2024diffusion, DBLP:journals/natmi/DuJGFHWDLSB24}.

In this section, we review recent advancements in molecule generation using flow matching techniques. We focus on methods that leverage continuous probability flow trajectories to generate novel molecular structures and properties, highlighting how flow matching has enhanced molecule generation.

\subsection*{2D Molecule Generation}
\begin{figure}[h]
    \centering
    \includegraphics[width=0.95\linewidth]{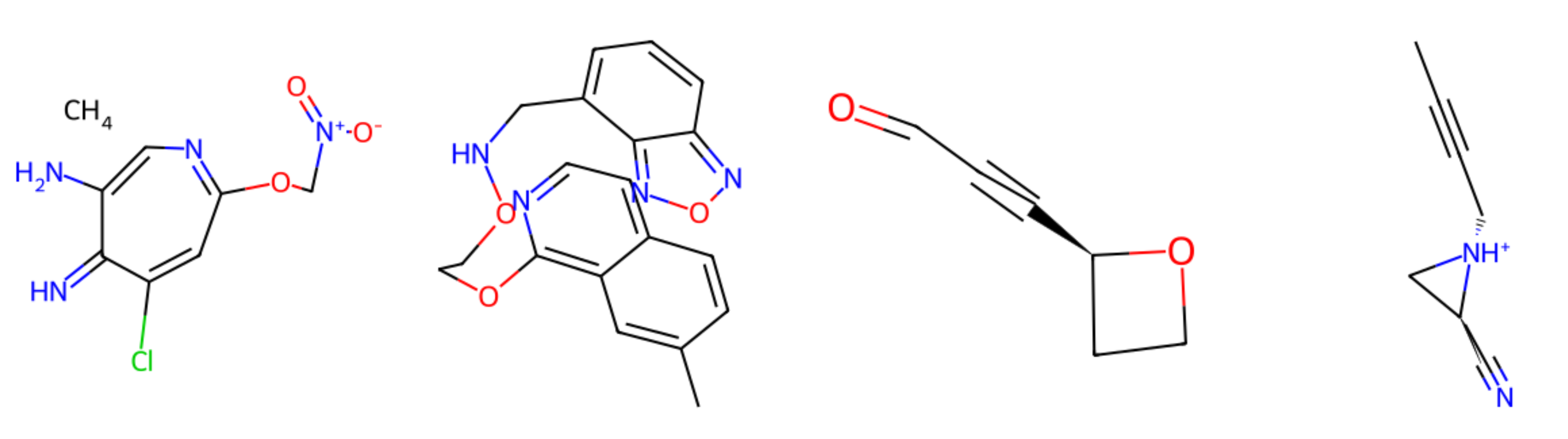}
    \caption{2D graph representations of example molecules generated from the GEOM-Drugs \cite{axelrod2022geom} (left two) and QM9 \cite{ramakrishnan2014quantum} (right two) datasets. Each molecule is visualized as a 2D graph, where atoms are nodes and chemical bonds are edges, capturing both structural and topological properties.}
    \label{fig:2d_example}
\end{figure}
Although real-world molecules are inherently three-dimensional objects, as illustrated in Figure \ref{fig:2d_example}, researchers often simplify the problem by using 2D graph-based molecular modeling when the 3D structure is not the primary focus \cite{DBLP:conf/ijcai/GuoGN0IMW0WZC23, de2018molgan, li2018multi}. This approach offers several advantages, including increased computational efficiency and reduced information requirements during inference.

Flow matching on graph data remains relatively unexplored, as the concept of flow matching itself is still under development. Nevertheless, existing studies often use 2D molecule generation as a preliminary test case to evaluate newly proposed flow matching variants. For instance, Eijkelboom et al. \cite{DBLP:conf/nips/EijkelboomBNWM24} combine flow matching with variational inference to introduce Variational Flow Matching for graph generation and CatFlow for handling categorical data. Additionally, GGFlow \cite{DBLP:journals/corr/abs-2411-05676} presents a discrete flow matching generative model that integrates optimal transport for molecular graphs. This model features an edge-augmented graph transformer, enabling direct communication among chemical bonds, thereby improving the representation of molecular structures. DeFoG \cite{qin2025defog} introduces a discrete formulation of flow matching tailored to the graph domain, explicitly decoupling the training and sampling phases to overcome inefficiencies in traditional diffusion-based models. By leveraging permutation-invariant graph matching objectives and exploring a broader sampling design space, DeFoG achieves strong empirical results on molecular graph generation with significantly fewer refinement steps.

\subsection*{3D Molecule Generation}
\setcounter{paragraph}{0}
\begin{figure}[h]
    \centering
    \includegraphics[width=0.95\linewidth]{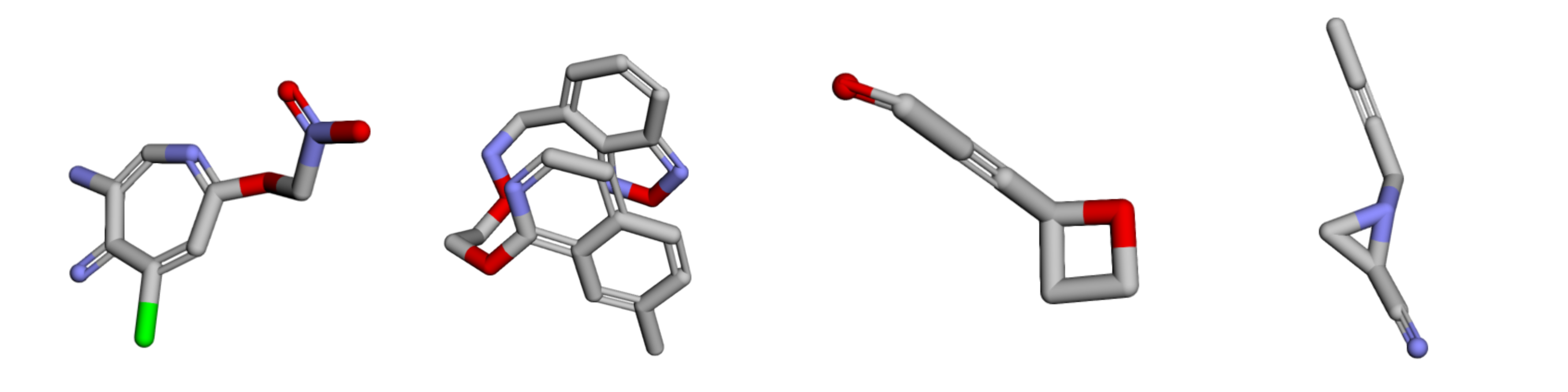}
    \caption{3D graph representations of example molecules generated from the GEOM-Drugs \cite{axelrod2022geom} (left two) and QM9 \cite{ramakrishnan2014quantum} (right two) datasets. Atoms are shown as nodes positioned in 3D Euclidean space, and bonds are represented as edges connecting them. These visualizations capture spatial geometry and stereochemistry important for molecular property prediction.}
    \label{fig:3d_example}
\end{figure}
Generating accurate 3D molecular structures is a critical task in drug discovery and structural biology \cite{baillif2023deep}. As illustrated in Figure \ref{fig:3d_example}, unlike 2D graph-based approaches, which primarily capture atomic connectivity, 3D molecular representations inherently encode spatial information, including bond angles, torsions, and stereochemistry. This spatial fidelity is essential for modeling interactions such as molecular docking, binding affinity, and conformational stability. While 2D representations cannot distinguish between stereoisomers or capture geometric nuances, 3D methods accurately model spatial conformation, enabling a more precise understanding of molecular properties \cite{DBLP:conf/icml/PengG0M23, hoogeboom2022equivariant, DBLP:conf/aaai/HuangZXW23}.

\paragraph{SE(3)-equivariant}
 To ensure physically meaningful and symmetry-consistent outputs, recent advancements have incorporated SE(3)-equivariant neural architectures into flow matching models. These models leverage the inherent symmetries of molecular systems, modeling graph generation as a continuous normalizing flow over node and edge features. For instance, Megalodon \cite{reidenbach2024applications} introduces scalable transformer models with basic equivariant layers, trained using a hybrid denoising objective to generate 3D molecules efficiently, achieving state-of-the-art results in both structure generation and energy benchmarks. EquiFM \cite{DBLP:conf/nips/SongGXCLEZM23} further improves the generation of 3D molecules by combining hybrid probability transport with optimal transport regularization, significantly speeding up sampling while maintaining stability. EquiFlow \cite{DBLP:journals/corr/abs-2412-11082} addresses the challenge of conformation prediction using conditional flow matching and an ODE solver for fast and accurate inference. By leveraging equivariant modeling, these methods improve the generation of valid and physically consistent molecular conformations, advancing the field of 3D molecule generation. Equivariant Variational Flow Matching \cite{equivariantvariationalfm} frames flow matching as a variational inference problem and enables both end-to-end conditional generation and post-hoc controlled sampling without retraining. The model further provides a principled equivariant formulation of VFM, ensuring invariance to rotations, translations, and atom permutations, which are essential for molecular applications.

\paragraph{Efficiency}

Generating high-quality 3D molecular structures efficiently is a major challenge in drug discovery and structural biology. While generative models have shown promise in modeling complex molecular structures, many existing approaches suffer from slow sampling speeds and computational inefficiency.
Flow matching-based methods leverage optimal transport and equivariant architectures to achieve faster and more reliable generation. For instance, GOAT \cite{hongaccelerating} formulates a geometric optimal transport objective to map multi-modal molecular features efficiently, using an equivariant representation space to achieve a double speedup compared to previous methods. MolFlow \cite{DBLP:journals/corr/abs-2406-07266} introduces scale optimal transport, significantly reducing sampling steps while maintaining high chemical validity. SemlaFlow \cite{irwin2025semlaflow} combines latent attention with equivariant flow matching, achieving an order-of-magnitude speedup with as few as 20 sampling steps. 
A recent work introduces SO(3)-Averaged Flow Matching with Reflow~\cite{caoefficient}, targeting both training and inference efficiency for 3D molecular conformer generation. The proposed SO(3)-averaged training objective leads to faster convergence and improved generalization compared to Kabsch-aligned or optimal transport baselines.
ET-Flow \cite{DBLP:conf/nips/HassanSLSTB24} leverages equivariant flow matching to generate low-energy molecular conformations efficiently, bypassing the need for complex geometric calculations. 
\paragraph{Guided Generation}
Guided and conditional generation enables the creation of structures that align with specific biological properties or conditions. In the context of flow matching, guided generation incorporates domain-specific knowledge to steer the generative process, while conditional generation aims to produce diverse outputs based on given inputs or contexts. These approaches are especially valuable in applications where accurate constraints are available.
Recent advancements in flow matching have introduced several methods to enhance guided and conditional generation. FlowDPO \cite{DBLP:conf/nips/JiaoK0024} addresses the challenge of 3D structure prediction by combining flow matching with Direct Preference Optimization (DPO), minimizing hallucinations while producing high-fidelity atomic structures. In conditional generation, Extended Flow Matching (EFM) \cite{DBLP:journals/corr/abs-2402-18839} generalizes the continuity equation, enabling more flexible modeling by incorporating inductive biases. 
For mixed-type molecular data, FlowMol \cite{DBLP:journals/corr/abs-2404-19739} extends flow matching to handle both continuous and categorical variables, achieving robust performance in 3D de novo molecule generation. 
3D energy-based flow matching \cite{zhouenergy2025} further enhances conditional generation by explicitly incorporating energy signals into both training and inference, improving structural plausibility and convergence. Together, these advances highlight the growing adaptability of flow-based approaches in generating biologically meaningful 3D molecular structures under domain constraints.
Additionally, OC-Flow \cite{DBLP:journals/corr/abs-2410-18070} leverages optimal control theory to guide flow matching without retraining, showing superior efficiency on complex geometric data, including protein design.

\subsection*{Conditional Molecule Design and Applications}
Recent advancements in flow matching for property-driven molecule design focus on not only generating the molecules themselves, but also predicting potential functionalities of the generated molecules. 
In scenarios requiring precise geometric control, GeoRCG \cite{DBLP:journals/corr/abs-2410-03655} enhances molecule generation by integrating geometric representation conditions, achieving significant quality improvements on challenging benchmarks.
Additionally, conditional generation with improved structural plausibility has been addressed by integrating distorted molecules into training datasets, as demonstrated in Improving Structural Plausibility in 3D Molecule Generation \cite{vost2024improving}. This method leverages property-conditioned training to selectively generate high-quality conformations. Stiefel Flow Matching \cite{DBLP:journals/corr/abs-2412-12540} tackles the problem of structure elucidation under moment constraints by embedding molecular point clouds within the Stiefel manifold, allowing for efficient and accurate generation of 3D structures with precise physical properties. Finally, IDFlow~\cite{zhouenergy} adopts an energy-based perspective on flow matching for molecular docking, where the generative process learns a deep mapping function to transform random molecular conformations into physically plausible protein-ligand binding structures. PropMolFlow \cite{zeng2025propmolflow} further advances property-guided molecule generation through a geometry-complete SE(3)-equivariant flow matching framework integrating five different property embedding methods with a Gaussian expansion of scalar properties. TemplateFM \cite{bergues2025template} introduces a ligand-based generation framework that leverages flow matching for template-guided 3D molecular alignment.

Structure-Based Drug Design (SBDD) is a key task in AI-assisted drug discovery, aiming to design small-molecule drugs that can bind to a given protein pocket structure. The main challenges in this domain lie in modeling the target protein structure, capturing protein–ligand interactions, enabling multimodal generation, and ensuring the chemical validity of generated molecules. In recent years, generative models have shown great potential in addressing these challenges, with Flow Matching (FM) models demonstrating unique advantages in multimodal modeling and generation efficiency. MolFORM \cite{huang2025molform} applies multimodal FM to the SBDD setting and employs DPO to optimize molecular binding affinity. FlexSBDD \cite{DBLP:conf/nips/ZhangWL24} further introduces protein pocket flexibility into the model, making it more reflective of real-world binding scenarios. In addition, MolCRAFT \cite{qumolcraft} adopts a Bayesian Flow Network (BFN) to model multimodal distributions in continuous parameter space, where BFN similarly defines a flow distribution. Moreover, \citep{xue2024unifying} reveals the equivalence between BFN, diffusion models, and stochastic differential equations (SDEs). PocketXMol \cite{peng2024decipher} provides a unified generative model for handling a variety of protein–ligand tasks. PAFlow \cite{zhou2025prior} introduces prior-guided flow matching with a learnable atom-number predictor to steer generation toward high-affinity regions and aligning molecule size with pocket geometry.

\section*{Protein Generation}\label{sec:protein}
\setcounter{paragraph}{0}

"Protein generation" can encompass a variety of tasks. To avoid confusion, we provide a brief comparison in Table \ref{tab:protein_tasks_comparison}.

\subsection*{Unconditional Generation}
\paragraph{Backbone Generation} 
Protein backbone generation aims to rapidly synthesize physically realizable 3D scaffolds that are diverse, designable, and functionally conditionable, while adhering to SE(3)-equivariance, local bond constraints, and global topological consistency. Recent efforts approach this challenge from two directions: enhancing the flow matching framework and improving protein feature representation learning. From the flow matching perspective, FrameFlow \cite{yim2023fast} accelerates diffusion by reframing it as deterministic SE(3) flow matching, cutting sampling steps five-fold and doubling designability over FrameDiff. Rosetta Fold diffusion 2 (RFdiffusion2) \cite{ahern2025atom} uses the RosettaFold All-Atom neural network architecture and is trained with flow matching for improved
training and generation efficiency. FoldFlow-SFM \cite{bose2024sestochastic} further extends this by introducing stochastic flows on SE(3) manifolds using Riemannian optimal transport, enabling the rapid generation of long backbones (up to 300 residues) with high novelty and diversity. Complementarily, recent work also advances architectural designs for protein representation learning. Yang et al. \citep{yan2025robust} combine global Invariant Point Attention (IPA) with local neighborhood aggregation to extract meaningful features, and further use ESMFold and AlphaFold3 to filter the invalid generated backbones. Wagner, Simon, et al. \cite{wagner2024generating} proposes Clifford frame attention (CFA), an extension of IPA by exploiting projective geometric algebra and higher-order message passing to capture residue-frame interactions, yielding highly designable proteins with richer fold topologies. FoldFlow-2 \cite{huguet2024sequence} augments SE(3) flows with PLM embeddings and a multi-modal fusion trunk, enabling sequence-conditioned generation with reinforced reward alignment and state-of-the-art diversity, novelty, and designability on million-scale synthetic–real datasets. Proteina \cite{geffner2025proteina} scales unconditional FM to a 400 M-parameter non-equivariant transformer trained on 21 M synthetic backbones, using hierarchical CATH conditioning to transport isotropic noise to native-like $C_\alpha$ traces. ProtComposer \cite{stark2025protcomposer} augments a Multiflow \cite{campbell2024generative} backbone with SE(3)-invariant cross-attention to user-sketched 3-D ellipsoid tokens, steering the FM vector field toward compositional spatial layouts while preserving unconditional diversity.

\paragraph{Co-design Generation}
Recent work reframes sequence–structure co-design as learning a unified vector field that jointly models discrete amino acid identities and continuous 3D coordinates, bypassing the traditional two-stage pipeline that separately samples a backbone before fitting a compatible sequence. This co-generative setting is especially challenging due to the need to reconcile fundamentally different data manifolds, enforce SE(3) symmetry, and ensure bidirectional invertibility, all while scaling to the vast combinatorial space of long proteins.
CoFlow \cite{yang2025co} proposes a joint discrete flow that models residue identities and inter-residue distances as CTMC states, augmented with a multimodal masked language module that allows structural flows and sequence tokens to condition each other. Discrete Flow Models (DFM) \cite{campbell2024generative} formalize flow matching on arbitrary discrete spaces by interpreting score-based guidance as CTMC generator reversal. Instantiated as MultiFlow, this framework enables sequence-only, structure-only, or joint generation within a single architecture-agnostic model, achieving state-of-the-art perplexity and TM-scores while being orders of magnitude faster than diffusion-based baselines. Finally, APM \cite{chen2025an} introduces a Seq\&BB module that jointly learns continuous SE(3) flows for backbone frames and discrete token flows for sequences, leveraging protein language models, Invariant Point Attention, and Transformer encoders to capture residue-level and pairwise interactions. APM supports precise interchain modeling and de novo design of protein complexes with specified binding properties.

\begin{table*}[t!]
\centering
\caption{Comparison of major protein modeling tasks. We highlight the distinctions in input, output, objective, and representative methods.}
\label{tab:protein_tasks_comparison}
\resizebox{0.99\textwidth}{!}{%
\begin{tabular}{llll}
\toprule
\textbf{Task} & \textbf{Input} & \textbf{Output} & \textbf{Objective} \\
\midrule
\textbf{Protein Structure Prediction} & 
Amino acid sequence & 
Full 3D structure (backbone + side chains) & 
Predict natural folded conformation \\

\textbf{Protein Design} & 
Target structure or functional constraint & 
Amino acid sequence (or full structure) & 
Design a sequence that folds into a desired structure or achieves a function \\

\textbf{Protein Backbone Generation} & 
Partial structure, constraints, or motifs & 
Backbone atomic coordinates (N, C$\alpha$, C) & 
Generate realistic backbone conformations as design templates \\
\bottomrule
\end{tabular}
}
\end{table*}

\subsection*{Conditional Generation}
\setcounter{paragraph}{0}

\paragraph{Motif-scaffolding Generation}
Motif-Scaffolding Generation. Conditional SE(3) flow-matching models embed fixed functional motifs into de-novo backbones by learning equivariant vector fields that respect both local motif geometry and global fold constraints, overcoming the diversity and fidelity limits of earlier diffusion approaches. FrameFlow-Motif \cite{yim2024improved} augments FrameFlow \cite{yim2023fast} with motif amortization and inference-time motif guidance, enabling scaffold generation around functional motifs with special-designed data augmentation and estimated conditional scores. EVA \cite{huangeva} casts scaffolding as geometric inverse design, steering a pretrained flow along motif-aligned probability paths to accelerate convergence and boost structural fidelity. RFdiffusion2 \cite{ahern2025atom} conducts catalytic site motif scaffolding at a much higher success rate, enabling de novo design of enzymes.

\paragraph{Pocket \& binder Design}

Conditional pocket and binder design tackles the dual challenge of sculpting a protein interface that both accommodates a specific ligand conformation and retains global fold stability, all while respecting SE(3) symmetry and the rich geometric-chemical priors that govern non-covalent recognition. Flow-matching models address these hurdles by learning equivariant vector fields that map an easy base distribution to the manifold of ligand-compatible protein–ligand complexes in a single, differentiable pass, avoiding the slow guidance loops and hand-crafted potentials of earlier diffusion or docking pipelines. AtomFlow \cite{liu2024design} unifies protein and ligand atoms into “biotokens” and applies atomic-resolution SE(3) flow matching to co-generate ligand conformations and binding backbones directly from a 2-D molecular graph. 
Additionally, FLOWR~\cite{cremer2025flowr} frames structure-aware ligand design as SE(3)-equivariant flow matching on a mixed continuous–categorical space. It learns the manifold of pocket-compatible molecules by coupling continuous FM for 3D atomic coordinates with categorical FM for fragment/chemotype identities, using equivariant optimal transport and an efficient pocket-conditioning mechanism to enforce interaction-aware constraints in a single pass.
Building on FLOWR~\cite{cremer2025flowr}, FLOWR.root~\cite{cremer2025flowrroot} unifies de novo generation, pharmacophore/interaction-conditional sampling, and fragment elaboration with joint heads for multi-endpoint affinity prediction and confidence estimation, sharing the conditional vector field while supervising downstream properties for multi-purpose structure-aware design.
FlowSite \cite{stark2024harmonic} introduces a self-conditioned harmonic flow objective that first aligns apo proteins to a harmonic potential and then co-generates discrete residue types and 3-D ligand poses, supporting multi-ligand docking and outperforming prior generative and physics-based baselines on pocket-level benchmarks. PocketFlow \cite{zhang2024generalized}  incorporates protein–ligand interaction priors (e.g., hydrogen-bond geometry) directly into the flow, then applies multi-granularity guidance to produce high-affinity pockets that significantly improve Vina scores and generalize across small molecules, peptides, and RNA ligands.
% \yanru{this one is protein and DNA, should exclude from molecule section?}
% \zh{may be protein - binding}
To efficiently recover all-atom structures from coarse-grained simulations, FlowBack \cite{DBLP:journals/jcisd/JonesKF25} utilizes flow matching to map coarse-grained representations to all-atom configurations, achieving high fidelity in protein and DNA structure reconstruction. 

\subsection*{Structure Prediction}
\setcounter{paragraph}{0}

\paragraph{Conformer Prediction}
Accurately sampling the conformational ensembles underlying protein function remains challenging due to the cost of exhaustive molecular dynamics. Recent work leverages sequence-conditioned, SE(3)-equivariant flow matching to efficiently generate diverse, physically consistent states aligned with experimental observables.
AlphaFold Meets Flow Matching \cite{jing2024alphafold} repurposes single-state predictors (AlphaFold, ESMFold) as generative engines by fine-tuning them under a harmonic flow-matching objective, yielding AlphaFlow/ESMFlow ensembles that surpass MSA-subsampled AlphaFold on the precision-diversity trade-off and reach equilibrium observables faster than replicate MD trajectories. 
P2DFlow \cite{jin2025p2dflow} augments SE(3) flow matching with a latent “ensemble” dimension and a physics-motivated prior, enabling it to reproduce crystallographic B-factor fluctuations and ATLAS MD distributions more faithfully than earlier baselines.

\paragraph{Side-chain Packing}

Predicting rotameric states for each residue requires joint compliance with steric constraints, energetic preferences, and SE(3)-equivariance. Recent work has explored constrained side-chain prediction through flow matching. FlowPacker \cite{lee2025flowpacker} formulates side-chain placement as torsional flow matching, coupling the learned vector field to EquiformerV2 \cite{liao2024equiformerv}, an SE(3)-equivariant graph attention backbone. PepFlow \cite{li2024full} generalizes this approach to full-atom peptides using a multi-modal flow that captures joint distributions over backbone frames, side-chain torsions, and residue identities. Partial sampling from this flow achieves state-of-the-art results in fixed-backbone packing and receptor-bound refinement, while maintaining full differentiability for downstream design applications.

\paragraph{Docking Prediction}

Recent work reframes protein-ligand docking as a flow-matching (FM) generative problem, replacing diffusion with a simulation-free objective that learns a bijective map from unbound receptors (apo) to bound complexes (holo). FlowSite \cite{stark2024harmonic} introduces a self-conditioned FM objective that harmonically couples translational, rotational and torsional degrees of freedom. By leveraging GAT and TFN layers for ligand–protein interaction modeling, it further extends to jointly generate contact residues and ligand coordinates, substantially improving sample quality, simplicity, and generality in pocket-level docking. Meanwhile, FlowDock \cite{morehead2025flowdock} learns a geometric flow mapping unbound to bound structures, while predicting per-complex confidence and binding affinity estimates. ForceFM~\cite{guoforcefm} reframes protein–ligand docking as force-guided manifold flow matching, injecting physics-based energy gradients into translational, rotational, and torsional flows to steer generation toward low-energy, physically realistic conformations.

\subsection*{Peptide and Antibody Generation}
Recent work \cite{li2024full, jin2025p2dflow, lin2024ppflow, huang2025non, kong2025protflow} formulates peptide design as conditional flow matching over multiple geometric and categorical manifolds, explicitly modeling residue type, spatial position, orientation, and angles in a unified generative framework. PepFlow \cite{li2024full} introduces the first multi-modal flow matching framework for protein structure design, jointly modeling residue positions via Euclidean CFM, orientations via Spherical CFM, angles via Toric CFM, and types via Simplex CFM. This unified approach achieves excellent performance on sequence recovery and side-chain packing in receptor-conditioned design tasks. D-Flow \cite{jin2025p2dflow} extends this paradigm to D-peptides by augmenting limited training data through a chirality-aware mirror transformation and incorporating a lightweight structural adapter into a pretrained protein language model. PPFlow \cite{lin2024ppflow} formulates peptide torsion generation as flow matching on a $(3n-3)$-torus with $n$ being the number of amino acids, while modeling global transitions and residue types via Euclidean flows and employing SO(3)-CFM for rotations. This formulation enables effective conditional sampling for diverse tasks such as peptide optimization and docking. Finally, NLFlow \cite{huang2025non} pioneers non-linear conditional vector fields by employing polynomial interpolation over the position manifold, enabling faster convergence toward binding pockets and effectively addressing temporal inconsistencies across modalities. This approach leads to improvements in structural stability and binding affinity compared to prior linear flow models. Collectively, these studies underscore the importance of manifold-specific flows, conditioning strategies, and geometric priors for scalable, high-fidelity peptide generation. In contrast to these geometry-intensive approaches, ProtFlow \cite{kong2025protflow} treats peptides as amino acid sequences and bypasses non-Euclidean representations by embedding each residue using a pretrained protein language model (PLM). In the embedding space of PLMs, ProtFlow trains a reflow-enabled sequence flow model that supports both single-step generation and multi-chain co-design. Collectively, these studies highlight the critical role of manifold-specific flows, conditioning strategies, and geometric priors in enabling scalable and high-fidelity peptide generation.

The study of antibody structure design with flow matching is emerging as well. For instance, FlowAB \cite{DBLP:conf/bibm/ZhangLBCLZ24} utilizes energy-guided SE(3) flow matching to improve antibody structure refinement, integrating physical priors to enhance CDR accuracy with minimal computational overhead. 

\section*{Other Bio Applications}\label{sec:other}

\begin{table*}[t]
\centering
\caption{Datasets and Software in Biology and Life Science to Test Flow Matching Methods (Part I)}
\resizebox{0.99\textwidth}{!}{%
\begin{tabular}{llccl}
\toprule
\textbf{Task} & \textbf{Dataset} & \textbf{Scale / Number of Samples} & \textbf{Links} & \textbf{Used By}\\
\midrule
\multirow{2}{*}{\makecell[c]{\textbf{DNA Sequence} \\ \textbf{Generation}}}
  & Promoter DNA Sequence   & 100, 000 & \href{https://arxiv.org/abs/2305.10699}{Paper1}; \href{https://arxiv.org/abs/2402.05841}{Paper2} \href{https://arxiv.org/abs/2405.14664}{Paper3}  \href{https://github.com/jzhoulab/ddsm}{Code1}; \href{https://github.com/HannesStark/dirichlet-flow-matching}{Code2}; \href{https://github.com/olsdavis/fisher-flow}{Code3};
  & \cite{stark2024dirichlet} \cite{davis2024fisher} \cite{tang2025gumbel}
  \\
  & Enhancer DNA Sequence   & 104,665 (fly brain); 88,870 (human melanoma)   & \href{https://arxiv.org/abs/2402.05841}{Paper1} \href{https://arxiv.org/abs/2405.14664}{Paper2}; \href{https://github.com/HannesStark/dirichlet-flow-matching}{Code1}; \href{https://github.com/olsdavis/fisher-flow}{Code2};
  & \cite{stark2024dirichlet} \cite{davis2024fisher} \\
\midrule
\multirow{3}{*}{\makecell[c]{\textbf{RNA Sequence} \\ \textbf{Generation}}}
    & Rfam Database \cite{griffithsjones2003rfam}
    & Over 20M sequences
    &  \href{https://pmc.ncbi.nlm.nih.gov/articles/PMC165453/}{Paper};
      \href{https://rfam.org/}{Homepage};   
      \href{https://huggingface.co/datasets/multimolecule/rfam}{Huggingface};       
    & \cite{gao2024rnacg} \\
& Muscle/PC3/HEK 5' UTR libraries \cite{chu2024utr}
    & 41,446 
    & \href{https://doi.org/10.1038/s42256-024-00823-9}{Paper};
    \href{https://codeocean.com/capsule/6711822}{Code}
    & \cite{gao2024rnacg} \\ 
& RNAsolo \cite{adamczyk2022rnasolo}
    & 18,808 RNA 3D structures
    & \href{https://academic.oup.com/bioinformatics/article/38/14/3668/6604270}{Paper};
    \href{https://rnasolo.cs.put.poznan.pl/}{Homepage}
    & \cite{nori2024rnaflow} \cite{rubin2025ribogen} \\ 
% \cmidrule{2-3}
%   & \textit{p-value}       & \textit{0.0492} \\
\midrule
\multirow{10}{*}{\makecell[c]{\textbf{Single-cell} \\ \textbf{Trajectory}}}
    & Pancreas single-cell data
    \cite{bastidas2019comprehensive}
    & 36,351 cells
    &  \href{https://journals.biologists.com/dev/article/146/12/dev173849/19483/Comprehensive-single-cell-mRNA-profiling-reveals-a}{Paper};
      \href{https://www.ncbi.nlm.nih.gov/geo/query/acc.cgi?acc=GSE132188}{Download Link}          
    & \cite{klein2024genot} \\
& Drug perturbation single-cell data
    \cite{srivatsan2020massively}
    & 650K single-cell transcriptomes 
    & \href{https://www.science.org/doi/10.1126/science.aax6234}{Paper};
    \href{https://github.com/bunnech/cellot/tree/main}{Download Instruction}
    & \cite{klein2024genot} \\ 
& Multi-modal single-cell analysis \cite{luecken2021sandbox}
    & 120K single cells (human bone marrow)
    & \href{https://datasets-benchmarks-proceedings.neurips.cc/paper_files/paper/2021/file/158f3069a435b314a80bdcb024f8e422-Paper-round2.pdf}{Paper};
    \href{https://openproblems.bio/results/predict_modality?version=v1.0.0}{Homepage}; \href{https://openproblems.bio/datasets}{Dataset List}
    & \cite{klein2024genot} \cite{palmamulti} \\ 
& PBMC
    \cite{derbois2023single}
    & 30K cells
    &  \href{https://www.nature.com/articles/s41597-023-02348-z}{Paper};
      \href{https://www.10xgenomics.com/cn/datasets/10-k-peripheral-blood-mononuclear-cells-pbm-cs-from-a-healthy-donor-single-indexed-3-1-standard-4-0-0}{Download Link}          
    & \cite{palma2024cellflow} \cite{palmamulti} \\
& Dentate gyrus dataset \cite{la2018rna}
    & 18,213 cells 
    & \href{https://www.nature.com/articles/s41586-018-0414-6}{Paper};
    \href{https://scvelo.readthedocs.io/en/stable/index.html}{scVelo Documentation}
    & \cite{palma2024cellflow} \cite{palmamulti} \\ 
& Human Lung cells Atlas \cite{sikkema2023integrated} 
    & 584,944. \cite{palma2024cellflow} uses a subset of 32,272 \cite{vieira2019cellular}
    & \href{https://www.nature.com/articles/s41591-023-02327-2}{Paper};
    \href{https://www.lungcellatlas.org/}{Homepage}; \href{https://openproblems.bio/datasets}{Dataset List}
    & \cite{palma2024cellflow} \cite{palmamulti} \\
& Tabula Muris \cite{schaum2018single} 
    & 245,389 cells
    & \href{https://www.nature.com/articles/s41586-018-0590-4}{Paper};
    \href{https://www.czbiohub.org/sf/tabula-muris/}{Homepage}; \href{https://github.com/czbiohub-sf/tabula-muris}{Code}
    &  \cite{palmamulti} \\
  & Embryoid Body (EB) \cite{moon2019visualizing}
    & 5 marginals
    & \href{https://doi.org/10.1038/s41587-019-0336-3}{Paper};
      \href{https://data.mendeley.com/datasets/v6n743h5ng/1}{Code} 
    & \cite{kapusniak2024metric} \\
  & CITE-seq (Cite) \cite{lance2022multimodal}
    & 4 marginals
    & \href{https://www.biorxiv.org/content/10.1101/2022.04.11.487796v1}{Paper};
      \href{https://openproblems.bio/datasets/openproblems_v1_multimodal/citeseq_cbmc}{Homepage}
    & \cite{kapusniak2024metric} \\
  & Multiome (Multi) \cite{lance2022multimodal}
    & 4 marginals
    & \href{https://www.biorxiv.org/content/10.1101/2022.04.11.487796v1}{Paper} 
      \href{https://www.10xgenomics.com/datasets/10-k-human-pbm-cs-multiome-v-1-0-chromium-x-1-standard-2-0-0}{Homepage} 

    & \cite{kapusniak2024metric} \\

\midrule
\multirow{11}{*}{\textbf{\makecell[c]{Molecule \\ Generation}}} 
  & \makecell[l]{Quantum Machine (QM) \cite{ramakrishnan2014quantum}}    
  & \makecell[c]{Various sizes. QM9: 133,885}
  & \makecell[c]{\href{https://www.nature.com/articles/sdata201422}{Paper};  \href{http://quantum-machine.org/datasets/}{Homepage}; \href{https://paperswithcode.com/dataset/qm9}{Paper With Code}; \href{https://www.kaggle.com/datasets/zaharch/quantum-machine-9-aka-qm9}{Kaggle}}
  & \makecell[l]{\cite{eijkelboom2024variational} \cite{qin2025defog} \cite{dunn2024mixed}  
    \cite{DBLP:journals/corr/abs-2411-05676} \cite{DBLP:conf/nips/EijkelboomBNWM24}  \\
    \cite{DBLP:conf/nips/SongGXCLEZM23} 
    \cite{DBLP:journals/corr/abs-2412-11082} \cite{reidenbach2024applications} 
    \cite{DBLP:conf/nips/SongGXCLEZM23} \cite{equivariantvariationalfm} 
    \cite{DBLP:journals/corr/abs-2406-07266} \\
    \cite{hongaccelerating} 
    \cite{DBLP:journals/corr/abs-2410-18070} \cite{DBLP:journals/corr/abs-2404-19739} 
    \cite{DBLP:journals/corr/abs-2410-03655} \cite{vost2024improving}
    \cite{DBLP:journals/corr/abs-2412-12540}
    } 
    
  \vspace{1mm} \\
  & \makecell[l]{ZINC \cite{irwin2012zinc}}          & \makecell[c]{Various sizes. ZINC250K: 249,456}  & \makecell[c]{\href{https://pubs.acs.org/doi/10.1021/ci3001277}{Paper}; \href{https://paperswithcode.com/dataset/zinc}{Paper With Code}; \href{https://huggingface.co/datasets/yairschiff/zinc250k}{Huggingface}; \href{https://www.kaggle.com/datasets/basu369victor/zinc250k}{Kaggle}} 
  & \makecell[l]{
  \cite{eijkelboom2024variational} \cite{qin2025defog}
  \cite{DBLP:journals/corr/abs-2411-05676}
  \cite{DBLP:conf/nips/EijkelboomBNWM24}\\
  \cite{DBLP:conf/nips/SongGXCLEZM23}
  \cite{DBLP:journals/corr/abs-2402-18839}
  \cite{equivariantvariationalfm} 
  \cite{vost2024improving}
  }
  \vspace{1mm} \\
  & Guacamol \cite{brown2019guacamol}           & 1,591,378  & \href{https://arxiv.org/abs/1811.09621}{Paper}; \href{https://github.com/BenevolentAI/guacamol}{Code}; \href{https://paperswithcode.com/paper/guacamol-benchmarking-models-for-de-novo}{Paper With Code}
  & \cite{qin2025defog}
  \cite{DBLP:conf/nips/SongGXCLEZM23}
  \\
  & MOSES \cite{polykovskiy2020molecular}           & 1,936,963  & \href{https://arxiv.org/abs/1811.12823}{Paper}; \href{https://github.com/molecularsets/moses}{Code}; \href{https://paperswithcode.com/dataset/moses}{Paper With Code}
  & \cite{qin2025defog}

  \cite{DBLP:conf/nips/SongGXCLEZM23}
  \vspace{1mm} \\
    & \makecell[l]{GEOM-Drugs \cite{axelrod2022geom}}  
    & \makecell[c]{430,000}
    & 
    \makecell[c]{\href{https://www.nature.com/articles/s41597-022-01288-4}{Paper};
    \href{https://github.com/learningmatter-mit/geom}{Code};
    \href{https://paperswithcode.com/dataset/geom-drugs}{Paper With Code}}
    &     \makecell[l]{\cite{dunn2024mixed}
    \cite{reidenbach2024applications}
    \cite{DBLP:conf/nips/SongGXCLEZM23} 
    \cite{equivariantvariationalfm}
    \cite{hongaccelerating}
    \cite{DBLP:journals/corr/abs-2406-07266} \\
    \cite{caoefficient}
    \cite{DBLP:journals/corr/abs-2404-19739}
    \cite{DBLP:conf/nips/HassanSLSTB24} 
    \cite{DBLP:journals/corr/abs-2410-03655}
    \cite{vost2024improving}
    \cite{DBLP:journals/corr/abs-2412-12540}
    } \vspace{1mm}
    \\
          & PoseBusters benchmark 
      \cite{buttenschoen2024posebusters}
    & 308 curated protein–ligand complexes  
    & \href{https://doi.org/10.1039/D3SC04185A}{Paper}; 
      \href{https://github.com/maabuu/Posebusters}{code}; 
      \href{https://paperswithcode.com/dataset/posebusters-benchmark}{Paper with code}  
    &  \cite{nagaraj2024igflow} \\
      & GEOM-QM9 \cite{axelrod2022geom}
    & 133,885
    & \href{https://www.nature.com/articles/s41597-022-01288-4}{Paper}; \href{https://github.com/learningmatter-mit/geom}{Code}; \href{https://paperswithcode.com/dataset/geom-qm9}{Paper With Code}
    & \cite{dunn2024mixed}
    \cite{DBLP:journals/corr/abs-2412-11082}
    \cite{caoefficient}
    \cite{DBLP:conf/nips/HassanSLSTB24}
    \\
    & SAbDab 
      \cite{dunbar2014sabdab}
    & $9{,}680$  
    & \href{https://doi.org/10.1093/nar/gkt1043}{Paper}; \href{https://opig.stats.ox.ac.uk/webapps/sabdab/}{Homepage}; 
    & \cite{DBLP:conf/nips/JiaoK0024} \cite{DBLP:conf/bibm/ZhangLBCLZ24} \cite{wu2025flowdesign}\\
\midrule
\multirow{1}{*}{\textbf{\makecell{Molecular Binder \\ Generation}}} 
  & Binding MOAD \cite{hu2005binding}
    & 41K complexes
    & \href{https://pubmed.ncbi.nlm.nih.gov/15971202/}{Paper}; \href{https://ngdc.cncb.ac.cn/databasecommons/database/id/65}{Homepage}
    & \cite{DBLP:conf/nips/ZhangWL24} \cite{stark2024harmonic} \cite{zhang2024generalized} \\
  & CrossDocked \cite{francoeur2020three}
    &  22.5M protein-molecule pairs
    & \href{https://pubs.acs.org/doi/10.1021/acs.jcim.0c00411}{Paper}; \href{https://github.com/gnina/models}{Code}
    & \cite{DBLP:conf/nips/ZhangWL24} \cite{zhang2024generalized} \\
\midrule
\multirow{1}{*}{\textbf{\makecell{Molecular Docking}}} 
  & PDBBind \cite{wang2004pdbbind}
    & 33,653 biomolecular complexes
    & \href{https://pubs.acs.org/doi/10.1021/jm030580l}{Paper}; \href{https://www.pdbbind-plus.org.cn/}{Homepage}
    & \cite{zhouenergy2025} \cite{stark2024harmonic} \cite{zhang2024generalized} \\
  & PPDBench \cite{agrawal2019benchmarking}
    & 133 protein-peptide complexes
    & \href{https://bmcbioinformatics.biomedcentral.com/articles/10.1186/s12859-018-2449-y}{Paper}; \href{https://webs.iiitd.edu.in/raghava/ppdbench/}{Homepage}
    & \cite{zhang2024generalized} \\
\midrule
\multirow{7}{*}{\makecell{\textbf{Protein Sequence} \\ \textbf{Design}}} 
  & UniRef \cite{suzek2015uniref}       & Various sizes. UniRef50: 70,198,728 & \href{https://academic.oup.com/bioinformatics/article/23/10/1282/197795}{Paper}; \href{https://www.uniprot.org/uniref}{Homepage}; \href{https://huggingface.co/datasets?search=uniref}{Huggingface}
  & \cite{cheng2025alpha} \cite{tang2025gumbel}  \\
  & Protein Data Bank (PDB) \cite{berman2000protein}       & Over 200K. 18,684 for curated version \cite{yim2023se} & \href{https://pmc.ncbi.nlm.nih.gov/articles/PMC102472/}{Paper}; \href{https://www.rcsb.org/}{Homepage}; \href{https://en.wikipedia.org/wiki/Protein_Data_Bank}{Wikipedia}
  & \cite{campbell2024generative} \cite{DBLP:journals/jcisd/JonesKF25} \cite{jones2025flowback} \\
  & Open Metagenomic Corpus (OMG) \cite{cornman2024omg}
    & 3.3B in total. OMG\_prot50: 207M
    & \href{https://www.biorxiv.org/content/10.1101/2024.08.14.607850v1}{Paper} ; \href{https://github.com/TattaBio/OMG}{Code}; \href{https://huggingface.co/datasets/tattabio/OMG}{HuggingFace}; \href{https://huggingface.co/tattabio/gLM2_650M}{Genomic LM}  
    & \cite{tang2025gumbel} \\
    & SAbDab
      \cite{dunbar2014sabdab}
    & $9{,}680$  
    & \href{https://doi.org/10.1093/nar/gkt1043}{Paper}; \href{https://opig.stats.ox.ac.uk/webapps/sabdab/}{Homepage}; 
    & \cite{nagaraj2024igflow} \cite{tan2025dyab} \\
      & OAS-paired antibody sequences  
      \cite{olsen2022observed}
    & $1.86\text{M pairs}$  
    & \href{https://doi.org/10.1002/pro.4205}{Paper}; \href{https://opig.stats.ox.ac.uk/webapps/oas/}{Homepage}
    & \cite{nagaraj2024igflow}  \\
      & RAbD Benchmark 
      \cite{adolf-bryfogle2018rosettaantibodydesign}
    & 60  
    & \href{https://doi.org/10.1371/journal.pcbi.1006112}{Paper}; \href{https://www.rosettacommons.org/docs/latest/application_documentation/antibody/RosettaAntibodyDesign}{Manual};  
    & \cite{tan2025dyab}\\
    & UniProt
      \cite{uniprot2018uniprot}
    & Over 60 million sequences
    & \href{https://academic.oup.com/nar/article/45/D1/D158/2605721}{Paper}; \href{https://www.uniprot.org/}{Homepage} 
    & \cite{kong2025protflow}
    \\
    & UniProtKB/SwissProt
      \cite{bairoch2000swiss}
    & 18364 sequence entries, 5,986,949 amino acids 
    & \href{https://pmc.ncbi.nlm.nih.gov/articles/PMC102476/}{Paper}; \href{https://www.expasy.org/resources/uniprotkb-swiss-prot}{Homepage} 
    & \cite{kong2025protflow}
    \\
\midrule
\multirow{3}{*}{\makecell{\textbf{Protein Backbone} \\ \textbf{Generation}}} 
  & Protein Data Bank (PDB) \cite{berman2000protein}       & Over 200K. 18,684 for curated version \cite{yim2023se} & \href{https://pmc.ncbi.nlm.nih.gov/articles/PMC102472/}{Paper}; \href{https://www.rcsb.org/}{Homepage}; \href{https://en.wikipedia.org/wiki/Protein_Data_Bank}{Wikipedia}
  & \cite{zhouenergy2025} \cite{yim2023fast} \cite{bose2024sestochastic} \cite{huguet2024sequence} \cite{liu2024design} \cite{yim2024improved} \\
  & SCOPe \cite{chandonia2022scope}       & 108,069  & \href{https://pubmed.ncbi.nlm.nih.gov/34850923/}{Paper}; \href{https://scop.berkeley.edu}{Homepage}
  & \cite{wagner2024generating} \cite{liu2024design} \\
  & Huguet et al. \cite{huguet2024sequence}       & 160K structures  & \href{https://arxiv.org/abs/2310.02391}{Paper}; \href{http://github.com/DreamFold/FoldFlow}{Code}
  & \cite{huguet2024sequence} \\
\midrule
\multirow{2}{*}{\textbf{\makecell{De Novo Protein \\ Generation}}} 
  & PepBDB \cite{wen2019pepbdb}
    & 13,299 peptide-protein complex
    & \href{https://academic.oup.com/bioinformatics/article/35/1/175/5050021}{Paper}; \href{http://huanglab.phys.hust.edu.cn/pepbdb/}{Homepage}; \href{https://github.com/MurtoHilali/PepBDB-ML}{PepBDB-ML}
    & \cite{lin2024ppflow} \\
  & PepMerge \cite{li2024full}
    & 8,365
    & \href{https://arxiv.org/abs/2406.00735}{Paper}; \href{https://github.com/Ced3-han/PepFlowww}{Code}
    & \cite{wu2024d} \cite{li2024full} \cite{huang2025non} \\
\midrule
\multirow{2}{*}{\makecell{\textbf{Protein Ensemble} \\ \textbf{Dynamics}}} 
  & Protein Data Bank (PDB) \cite{berman2000protein}       & Over 200K proteins & \href{https://pmc.ncbi.nlm.nih.gov/articles/PMC102472/}{Paper}; \href{https://www.rcsb.org/}{Homepage}; \href{https://en.wikipedia.org/wiki/Protein_Data_Bank}{Wikipedia}
  & \cite{jing2024alphafold}\\
  & ATLAS \cite{vander2024atlas}       & 1390 protein chains \cite{yim2023se} & \href{https://academic.oup.com/nar/article/52/D1/D384/7438909}{Paper}; \href{https://www.dsimb.inserm.fr/ATLAS}{Homepage}
  & \cite{jing2024alphafold} \cite{jin2025p2dflow} \\
\midrule
\makecell{\textbf{Protein Docking or} \\ \textbf{Side-chain Packing}}
  & \makecell[l]{CASP \cite{kryshtafovych2023critical}}       & \makecell[c]{Various sizes} & \makecell[c]{\href{https://pmc.ncbi.nlm.nih.gov/articles/PMC102472/}{Paper}; \href{https://predictioncenter.org/casp16/index.cgi}{Homepage}}
  & \makecell[l]{\cite{lee2025flowpacker} \cite{morehead2025flowdock}} \\
\midrule
\multirow{4}{*}{\textbf{\makecell{Peptide Binder \\ Design}}} 
  % & PDBBind (protein-ligand complexes) \cite{wang2004pdbbind}
  & PDBBind \cite{wang2004pdbbind}
    & 33,653 biomolecular complexes
    & \href{https://pubs.acs.org/doi/10.1021/jm030580l}{Paper}; \href{https://www.pdbbind-plus.org.cn/}{Homepage}
    & \cite{nori2024rnaflow} \cite{yan2025robust} \cite{morehead2025protein} \\
  & PepNN (peptide binding sites) \cite{abdin2022pepnn}
    & Varoius sizes ranging from 251 to 2,517  
    & \href{https://doi.org/10.1038/s42003-022-03445-2}{Paper}; \href{https://gitlab.com/oabdin/pepnn}{Code}; \href{https://en.wikipedia.org/wiki/PDBbind_database}{Wikipedia}  
    & \cite{tang2025gumbel} \\
    
  & BioLip2 \cite{zhang2024biolip2} 
    & 385,160 protein chains; 781,684 interactions
    & \href{https://academic.oup.com/nar/article/52/D1/D404/7233921?login=false}{Paper}; \href{https://zhanggroup.org/BioLiP/}{Homepage}; \href{https://github.com/kad-ecoli/mmCIF2BioLiP}{Web Code}  
    & \cite{tang2025gumbel} \\
  & Binder discrimination dataset \cite{nagaraj2024igflow} 
    & 4,883 antibody–antigen complexes  
    & \href{https://doi.org/10.1101/2025.05.06.652551}{Paper};  
    & \cite{nagaraj2024igflow}  \\
\midrule

\multirow{2}{*}{\textbf{\makecell{Peptide Design}}} 
  & PepBDB \cite{wen2019pepbdb}
    & 13,299 peptide-protein complex
    & \href{https://academic.oup.com/bioinformatics/article/35/1/175/5050021}{Paper}; \href{http://huanglab.phys.hust.edu.cn/pepbdb/}{Homepage}; \href{https://github.com/MurtoHilali/PepBDB-ML}{PepBDB-ML}
    & \cite{lin2024ppflow} \\
  & PepMerge \cite{li2024full}
    & 8,365
    & \href{https://arxiv.org/abs/2406.00735}{Paper}; \href{https://github.com/Ced3-han/PepFlowww}{Code}
    & \cite{wu2024d} \cite{li2024full} \cite{huang2025non} \\

\bottomrule

\end{tabular}
}
\label{tab:datasets}
\end{table*}

\begin{table*}[t]
\centering
\caption{Datasets and Software in Biology and Life Science to Test Flow Matching Methods (Part II)}
\resizebox{0.99\textwidth}{!}{%
\begin{tabular}{llccl}
\toprule
\textbf{Task} & \textbf{Dataset} & \textbf{Scale / Number of Samples} & \textbf{Links} & \textbf{Used By}\\
\midrule

\multirow{3}{*}{\makecell[c]{\textbf{Cell Morphology} \\ \textbf{Profiling}}}
  & BBBC021 \cite{ljosa2012annotated}
    & 39,600 images
    & 
    \href{https://doi.org/10.1038/nmeth.2083}{Paper};
    \href{https://bbbc.broadinstitute.org/BBBC021}{Homepage}; 
    & \cite{zhang2025cellflux} \\
  & RxRx1  \cite{taylor2019rxrx1}
    & 125,510 images
    & \href{https://aiforsocialgood.github.io/iclr2019/accepted/track1/pdfs/30_aisg_iclr2019.pdf}{Paper};
    
    \href{https://www.rxrx.ai/datasets}{Homepage}; \href{https://github.com/recursionpharma/rxrx-datasets/tree/trunk/rxrx1}{Code}; \href{https://paperswithcode.com/dataset/rxrx1}{Paper With Code}
    & \cite{zhang2025cellflux} \\
  & JUMP Cell Painting \cite{chandrasekaran2023jump}
    & 1.6 billion  profiles
    & \href{https://doi.org/10.1101/2023.03.23.534023}{Paper};  \href{https://github.com/gwatkinson/jump_download}{Code};
    \href{https://registry.opendata.aws/cellpainting-gallery/}{AWS}
    & \cite{zhang2025cellflux} \\
\midrule
\multirow{4}{*}{\makecell[c]{\textbf{Medical Image} \\ \textbf{Segmentation}}}
  & MoNuSeg \cite{Kumar2020MoNuSeg}
    & 30 train + 14 test images 
    
    & 
    \href{https://doi.org/10.1109/TMI.2019.2947628}{Paper};
    \href{https://monuseg.grand-challenge.org/Data/}{Homepage};

    & \cite{DBLP:journals/corr/abs-2405-18087}  \\
  & GlaS \cite{Sirinukunwattana2017GlaS}
    & 85 train + 80 test images
    & 
    \href{https://doi.org/10.1016/j.media.2016.08.008}{Paper};
    \href{https://warwick.ac.uk/fac/cross_fac/tia/data/glascontest/}{Homepage}; 
      
    & \cite{DBLP:journals/corr/abs-2405-18087}  \\
  & CAMUS  \cite{leclerc2019deep}
    & 450 patients; 1,600 images 
    & \href{https://doi.org/10.1109/TMI.2019.2900516}{Paper}; 
      \href{https://www.creatis.insa-lyon.fr/Challenge/camus/}{Homepage} 
    & \cite{DBLP:journals/corr/abs-2503-00266} \\
  & MSD Brain MRI \cite{antonelli2022medical}
    & 750 scans (T1-weighted) 
    & \href{https://doi.org/10.1038/s41467-022-30695-9}{Paper}; 
            \href{https://registry.opendata.aws/msd/}{AWS};
      \href{https://paperswithcode.com/dataset/medical-segmentation-decathlon}{PapersWithCode} 
    & \cite{DBLP:journals/corr/abs-2503-00266} \\
\midrule
\multirow{2}{*}{\makecell[c]{\textbf{MRI} \\ \textbf{Reconstruction}}}
  & fastMRI  \cite{Knoll2020fastMRI}
    & Knee: 1,398 scans; Brain: 7,002 scans
    & \href{https://fastmri.med.nyu.edu/}{Homepage}; 
      \href{https://doi.org/10.1148/ryai.2020190007}{Paper}; 
      \href{https://github.com/facebookresearch/fastMRI}{Code} 
    & \cite{DBLP:journals/cbm/ZhangHXD24} \\
  & BraTS-2020 (Brain Tumor Segmentation) \cite{Menze2015BRATS}
    & 494 subjects , 240×240 
    & \href{https://www.med.upenn.edu/cbica/brats2020/data.html}{Homepage}; 
      \href{https://doi.org/10.1109/TMI.2014.2377694}{Paper}; 
      \href{https://www.kaggle.com/datasets/awsaf49/brats2020-training-data}{Kaggle} 
    & \cite{DBLP:journals/cbm/ZhangHXD24}\\
\midrule
\multirow{2}{*}{\makecell[c]{\textbf{Spatial} \\ \textbf{Transcriptomics}}}
  & HEST-1k \cite{jaume2024hest}
    & 1,229 profiles 
    & \href{https://arxiv.org/abs/2406.16192}{Paper}; 
      \href{https://github.com/mahmoodlab/hest}{Code}
    & \cite{huang2025scalable} \\
  & STImage-1K4M \cite{chen2024stimage}
    & 1,149 slides (4 293 195 spots) 
    & \href{https://arxiv.org/abs/2406.06393}{Paper}; 
      \href{https://github.com/JiawenChenn/STimage-1K4M}{Code}; 
      \href{https://huggingface.co/datasets/jiawennnn/STimage-1K4M}{HuggingFace}
    & \cite{huang2025scalable} \\
\midrule
\multirow{2}{*}{\makecell[c]{\textbf{Single-cell} \\ \textbf{Omics}}}
  & seqFISH \cite{Lohoff2022Integration}
    & 351 genes; 29 cells per niche
    & \href{https://doi.org/10.1038/s41587-021-01006-2}{Paper};
      \href{https://doi.org/10.18129/B9.bioc.MouseGastrulationData}{Code}
    & \cite{haviv2025wasserstein} \\
  & scRNA-seq \cite{Stephenson2021COVIDMultiOmics}
    & 32 PCs per meta-cell
    & \href{https://doi.org/10.1038/s41591-021-01329-2}{Paper};
      \href{https://singlecell.broadinstitute.org/single_cell/study/SCP1107/single-cell-multi-omics-analysis-of-the-immune-response-in-covid-19}{Code}
    & \cite{haviv2025wasserstein} \\
  \midrule
\multirow{2}{*}{\makecell[c]{\textbf{Neural} \\ \textbf{Time Series}}}
  & Mouse brain LFP  \cite{steinmetz2019distributed}
    & 50 marginals (50–500 ms)
    & \href{https://doi.org/10.1038/s41586-019-1787-x}{Paper};  
      \href{https://figshare.com/articles/dataset/LFP_data_from_Steinmetz_et_al_2019/9727895}{Code}
    & \cite{wei2025streamlevel} 
    
    \\ \\
    \midrule
\multirow{3}{*}{\makecell[c]{\textbf{Neural Population}\\\textbf{Dynamics}}}
  & CO-C (Monkey C) \cite{Flint2012DREAM}
    & 5 sessions; 957 units
    & \href{https://crcns.org/data-sets/movements/dream}{DREAM}; 
      \href{https://doi.org/10.1088/1741-2560/9/5/056009}{Paper}
    & \cite{wang2025flow} \\
  & CO-M (Monkey M) \cite{Churchland2012NeuralDynamics}
    & 9 sessions; 1,728 units
    & \href{https://crcns.org/data-sets/motor-cortex/pmd-1}{pmd-1}; 
      \href{https://doi.org/10.1038/nature11129}{Paper}
    & \cite{wang2025flow} \\
  & RT-M (Monkey M) \cite{Cornblath2021NLB}
    & 1 session; 130 units
    & \href{https://neurallatents.github.io/datasets.html}{NLB-RTT}; 
      \href{https://neurallatents.github.io/assets/CornblathHeraviCunninghamSussillo_NeurIPS21.pdf}{Paper}
    & \cite{wang2025flow} \\

\bottomrule

\end{tabular}
}
\label{tab:datasets_II}
\end{table*}

\subsection*{Dynamic Cell Trajectory Prediction}

Dynamic Cell Trajectory. Generative trajectory models seek to reconstruct the continuously branching, stochastic evolution of cells from high-dimensional, sparsely sampled single-cell readouts, which is an endeavor hampered by severe noise, irregular time points, and the risk that straight Euclidean interpolants stray outside the biological manifold. CellFlow \cite{zhang2025cellflux} tackles this by framing morphology evolution under perturbations as an image-level flow-matching problem on cellular masks, enabling realistic, perturbation-conditioned movies of shape change that outperform diffusion and GAN baselines in both faithfulness and diversity. GENOT-L \cite{klein2024genot} introduces an entropic Gromov-Wasserstein flow that couples gene-expression geometry across time points, producing probabilistic lineage trajectories that capture heterogeneity and branching better than optimal-transport predecessors while remaining simulation-free. Metric Flow Matching \cite{kapusniak2024metric} instead learns geodesic vector fields under a data-induced Riemannian metric, yielding smoother interpolations that respect the manifold’s curvature and achieving state-of-the-art accuracy on single-cell trajectory benchmarks with fewer artifacts than Euclidean flows. Diversified Flow Matching \cite{diversifiedfm} extends this line of work by ensuring translation identifiability across diverse conditional distributions, a key challenge in modeling heterogeneous cellular states. Unlike prior GAN-based solutions, this work formulates the problem within an ODE-based flow matching framework, offering stable training and explicit transport trajectories. Collectively, these works highlight the importance of geometry-aware objectives and probabilistic conditioning for faithful dynamic cell-state generation.

\subsection*{Bio-image Generation and Enhancement}
Leveraging continuous probability flow to efficiently model biological structures, flow matching has shown great potential for bio-image generation and enhancement, enabling faster and more accurate modeling of complex biological data.
One notable application is FlowSDF \cite{DBLP:journals/corr/abs-2405-18087}, which introduces image-guided conditional flow matching for medical image segmentation. By modeling signed distance functions (SDF) instead of binary masks, FlowSDF achieves smoother and more accurate segmentation. This method also generates uncertainty maps, enhancing robustness in prediction tasks.
For medical image synthesis, an optimal transport flow matching approach \cite{DBLP:journals/corr/abs-2503-00266} addresses the challenge of balancing generation speed and image quality. By creating a more direct mapping between distributions, this method reduces inference time while maintaining high-quality outputs, and supports diverse imaging modalities, including 2D and 3D.
In MR image reconstruction, Multi-Modal Straight Flow Matching (MMSFlow) \cite{DBLP:journals/cbm/ZhangHXD24} significantly reduces the number of inference steps by forming a linear path between undersampled and reconstructed images. Leveraging multi-modal information with low- and high-frequency fusion layers, MMSFlow achieves state-of-the-art performance in fastMRI and Brats-2020 benchmarks.

\subsection*{Cellular Microenvironments from Spatial Transcriptomics}

Flow matching has also emerged as a powerful framework for modeling spatial transcriptomics (ST) data, which captures gene expression levels across spatial locations within a tissue. The core task in ST involves reconstructing or generating spatially-resolved gene expression maps that reflect underlying cellular microenvironments and tissue organization.
One such method is STFlow \cite{huang2025scalable} which introduces a scalable flow matching framework for generating spatial transcriptomics data from whole-slide histology images. It models the joint distribution of gene expression across all spatial spots in a slide, thereby explicitly capturing cell-cell interactions and tissue organization.
Complementarily, Wasserstein Flow Matching (WFM) \cite{haviv2025wasserstein} generalizes flow-based generative modeling to families of distributions. It introduces a principled way to model both 2D and 3D spatial structures of cellular microenvironments, and leverages the geometry of Wasserstein space to better match distributional characteristics across biological contexts.
Together, these methods highlight the utility of flow matching in capturing the spatially-aware, high-dimensional distributions characteristic of modern transcriptomics datasets.

\subsection*{Neural Activities}
Flow matching has recently shown promise in modeling and aligning neural activity, particularly for time series and brain-computer interface (BCI) applications, where neural signals are often stochastic and nonstationary.
Stream-level Flow Matching with Gaussian Processes \cite{wei2025streamlevel} extends conditional flow matching by introducing streams, which arelatent stochastic paths modeled with Gaussian processes. This reduces variance in vector field estimation, enabling more accurate modeling of correlated time series such as neural recordings.
Flow-Based Distribution Alignment \cite{wang2025flow} tackles inter-day neural signal shifts in BCIs through source-free domain adaptation. By learning stable latent dynamics via flow matching and ensuring stability through Lyapunov analysis, it enables reliable few-trial neural adaptation across days.
These approaches highlight the versatility of flow matching for neural data, supporting both high-fidelity generation and robust adaptation with limited supervision.
DIFFEOCFM \cite{collas2025riemannian} introduces Riemannian flow matching for brain connectivity matrices by leveraging pullback metrics to perform conditional FM on matrix manifolds, enabling efficient vector-field learning and fast sampling while preserving manifold constraints.

\section*{Evaluation Tasks and Datasets}
\label{sec: datasets}

In this section, we summarize evaluation tasks and datasets used for assessing flow matching methods in biology and life sciences. As listed in Table \ref{tab:datasets} and Table \ref{tab:datasets_II}, these tasks span a wide spectrum of domains, including genomics, transcriptomics, molecular chemistry, and structural biology. For each dataset, we also report its data scale or number of samples. Flow matching has been applied to a diverse set of generation and modeling problems, such as biological sequence generation, cell trajectory inference, molecule design, and protein structure modeling.

\textit{Sequence-Level Generation.} Flow matching models have been evaluated on tasks like DNA \cite{stark2024dirichlet,davis2024fisher,tang2025gumbel}, RNA \cite{griffithsjones2003rfam,adamczyk2022rnasolo,chu2024utr}, and protein \cite{suzek2015uniref,cornman2024omg,berman2000protein} sequence generation. These datasets range from promoter and enhancer sequences to large-scale protein and metagenomic corpora, covering both canonical and noncoding regions of the genome.

\textit{Single-Cell Modeling and Trajectory Inference.} Flow matching has been used to model temporal or conditional transitions in high-dimensional single-cell gene expression data, including developmental trajectories \cite{bastidas2019comprehensive}, perturbation responses \cite{srivatsan2020massively}, and modality prediction \cite{luecken2021sandbox}. Datasets such as PBMC \cite{derbois2023single}, dentate gyrus~\cite{la2018rna}, and Tabula Muris \cite{schaum2018single} provide diverse experimental contexts for evaluating these tasks.

\textit{Molecular Generation and Conformation Modeling.} Datasets such as QM9 \cite{ramakrishnan2014quantum}, ZINC \cite{irwin2012zinc}, GEOM-Drugs \cite{axelrod2022geom}, and MOSES \cite{polykovskiy2020molecular} provide chemically diverse molecular structures, enabling evaluation of molecular validity, novelty, and 3D geometry. Flow matching models are tested on their ability to generate, edit, or align molecular graphs and conformers.

\textit{Protein and Complex Design.} Structural datasets like SCOPe \cite{chandonia2022scope}, ATLAS \cite{vander2024atlas}, and curated PDB subsets support evaluation of flow-based models on protein backbone generation, folding, and structural refinement. Complementary datasets such as Binding MOAD \cite{hu2005binding}, CrossDocked \cite{francoeur2020three}, BioLip2 \cite{zhang2024biolip2}, and PepBDB \cite{wen2019pepbdb} enable studies on molecular docking, peptide-protein interactions, and binder generation.

Notably, many datasets are reused across different tasks due to their structural richness and biological relevance. For instance, the Protein Data Bank (PDB) \cite{berman2000protein} is used in tasks ranging from protein sequence design and backbone generation to modeling conformational dynamics and performing docking. Similarly, SAbDab \cite{dunbar2014sabdab} supports antibody sequence generation, structural modeling, and binder discrimination.

Despite the growing adoption of flow matching in biology, the field still lacks unified benchmarks for many tasks. This is likely due to the inherent heterogeneity of biological problems, ranging from sequence to structure, from single-cell to population scale, which makes standardized evaluation more challenging.  This stands in contrast to fields like computer vision or NLP, where well-defined benchmarks are more prevalent \cite{chang2024survey, wang2022generalizing, deng2009imagenet, dwivedi2023benchmarking}. Continued efforts in dataset curation and task formulation are needed to support consistent and reproducible assessment of generative models in the life sciences.

\section*{Futrue Direction}
\subsection*{Flow Matching for Discrete Sequence Generation}

Flow Matching has recently emerged as a promising generative modeling paradigm, offering a compelling balance between generation quality and training stability. While its success in continuous domains like image and molecule generation has been widely documented, applying FM to discrete sequence generation—especially in domains such as natural language, genomics, and code—remains a vibrant and largely underexplored frontier.

One of the most intriguing directions lies in understanding the representational advantages of discrete Flow Matching compared to traditional paradigms such as Masked Language Modeling (MLM). Unlike MLM, which relies on partial observation and token masking, FM provides a direct mapping from a base distribution to the target sequence via a continuous probability flow. This raises the question: Can discrete FM yield more semantically coherent representations and facilitate better downstream performance in tasks such as classification? Recent advances, such as Fisher Flow \cite{davis2024fisher} and Dirichlet FM \cite{stark2024dirichlet}, demonstrate that geometry-aware formulations over the probability simplex can encode meaningful geometric constraints and structure-aware trajectories, enabling more faithful modeling of discrete data distributions.

Another fundamental question concerns the generation capabilities of discrete FM relative to autoregressive (AR) models. While AR models remain the gold standard in natural language generation due to their strong likelihood modeling and contextual fluency, they suffer from slow sampling and exposure bias. In contrast, discrete FM supports parallel generation through ODE integration or sampling over learned Markov trajectories, offering substantial efficiency gains. However, its generation quality still lags behind state-of-the-art AR transformers in language generation \cite{davis2024fisher}, prompting future research into architectural refinements and better training objectives. 

Furthermore, the integration of FM with Transformer architectures remains an open challenge. Existing Transformer-based FM models either operate in latent embedding space or use discrete-continuous relaxations (e.g., Gumbel-Softmax) to approximate gradient flows. Yet, the Transformer’s causal attention structure may be suboptimal for non-autoregressive FM-based sequence generation, especially in domains where left-to-right order is arbitrary or non-existent (e.g., protein sequences, biological pathways). This invites research into order-agnostic architectures or the use of permutation-invariant encoders to better align with FM-based modeling.

Finally, flow matching may offer unique advantages in non-language sequence modeling tasks, such as biomolecular design and genome modeling, where biological constraints (e.g., base-pairing, structural motifs) must be enforced. Unlike language, these sequences often lack natural generation order and exhibit rich multi-modal dependencies. FM’s ability to incorporate conditioning, geometry-aware constraints, and structure-guided generation (e.g., via SE(3)-equivariant or manifold-aware flows) makes it a particularly attractive candidate. Future work may focus on developing discrete FM formulations that are not only domain-adaptive, but also biologically interpretable and sample-efficient.

\subsection*{Small Molecule Generation and Modeling}
Small molecule generation is a core task in cheminformatics and drug discovery, where FM has recently shown promising capabilities in both unconditional and conditional generation settings. By modeling continuous probability flows between simple priors and molecular distributions, FM offers an appealing alternative to diffusion models, with improved sample efficiency and the potential to integrate domain knowledge. However, due to the scarcity of molecular structure data and the complexity of structural constraints, several key challenges remain before FM can fully realize its potential for small molecule generation.

One fundamental limitation lies in the data scarcity and structural heterogeneity of small molecule datasets. Unlike macromolecules such as proteins, which benefit from large-scale structural repositories (e.g., PDB), small molecule datasets are often limited in size and diversity, especially for annotated 3D conformers. As a result, FM models trained on these datasets may struggle to generalize across different chemical scaffolds, limiting their utility in low-resource or out-of-distribution scenarios. Addressing this issue may require more effective data augmentation strategies (e.g., using force field simulations or generative conformer expansion), transfer learning pipelines, or semi-supervised flow matching objectives that make better use of unlabeled data.

To improve the physical plausibility and functional relevance of generated small molecules, a key direction lies in incorporating domain-specific inductive priors into both the training and sampling stages of flow matching. Small molecules are governed by well-defined chemical and physical constraints—such as bond lengths and angles, valence rules, charge distributions, and conformational energetics—which can be explicitly modeled to constrain the learned probability flow. Embedding such priors into the vector field design or generation trajectories (e.g., via energy-guided loss functions or structure-aware conditioning) can substantially improve the realism and synthesizability of generated compounds.

At the same time, enhancing the conditional generation capabilities of FM is essential for tasks that demand goal-directed molecular design, such as generating molecules with desired pharmacological properties, satisfying functional group templates, or fitting into predefined binding pockets. Conditional flow matching offers a natural framework for structure- and property-guided generation, enabling fine-grained control over outputs via learned trajectories that satisfy specific constraints. Future work may explore more expressive conditioning schemes, multi-property guidance, or interaction-aware control mechanisms, paving the way for FM-based models to support precision molecular design in high-stakes domains such as drug discovery and materials engineering.

A further challenge lies in modeling molecular interactions and dynamic processes. Molecular docking and binding affinity prediction remain critical tasks in early-stage drug design, requiring models to account for conformational flexibility in small molecules and the adaptive nature of protein binding pockets, particularly with respect to side-chain rearrangements. Even more challenging tasks, such as enzyme design, involve not just molecular recognition but also modeling of specific reaction mechanisms. Thus, leveraging the FM framework to capture inter-molecular interactions and reaction dynamics represents a crucial and promising direction for future research.

\subsection*{Protein}

In the field of protein modeling, Flow Matching (FM) has emerged as an efficient approach for sequence and structure modeling, demonstrating complementary advantages to traditional methods. Proteins, as highly complex biological macromolecules, exhibit a unique combination of discrete primary sequences and continuous three-dimensional structures, which poses distinct challenges for the design and training of FM-based models.

One important future direction is to establish effective matching mechanisms across different protein modalities. For example, in mapping from amino acid sequences to 3D structures, FM could serve as a bridge between discrete and continuous spaces, enhancing the model’s expressiveness in structure prediction and generation tasks. Furthermore, in applications such as protein-protein docking and complex assembly modeling, FM offers a promising framework for capturing transformation paths in high-dimensional, complex spaces.

In addition, modeling protein dynamics—such as conformational changes or ligand-induced fit—remains a core challenge in structural biology. Future work may explore integrating FM with physical simulations (e.g., molecular dynamics) or diffusion-based processes, enabling the learning of natural transition paths between protein states and improving interpretability of their functional mechanisms.
\section*{Conclusion}
\label{sec: conclusion}

Flow matching has become a compelling alternative to diffusion-based generative modeling, offering advantages in stability, efficiency, and control. In this survey, we provide a structured overview of its growing use in biology and life sciences, covering a diverse range of tasks from sequence generation and molecular design to protein modeling. We also compile a comprehensive list of datasets used for evaluation, including their scale and cross-task applicability.
Despite promising progress, we also summarize the challenges that the field faces. We hope this survey could clarify current trends and motivate future research at the intersection of generative modeling and the life sciences.

\section*{Declarations}
\subsection*{Author Contributions Statement}
Zihao L, ZZ, and XL drafted the main manuscript text. FF reviewed the datasets and benchmarks and assisted with publishing the GitHub resources repository. YQ and ZX provided oversight of the biology-related and flow-matching-related sections, respectively, and contributed extensive feedback. Zhining L prepared the figures and contributed to Section "Other Bio Applications". XN and TW contributed to Sections "Challenges of Generative Modeling for Biology", "Connection to Existing Survey", and "Conclusion". Dr. GL, Dr. HT, and Dr. JH supervised the research. All authors reviewed the manuscript and provided valuable suggestions. All authors have read and approved the manuscript.

\subsection*{Competing Interest Statement}
The corresponding author, Dr.Jingrui He, serves as an Associate Editor for \textit{npj Artificial Intelligence}. Aside from this editorial role, the authors declare no other competing financial or non-financial interests as defined by Nature Portfolio, or other interests that might be perceived to influence the results and/or discussion reported in this paper.

\subsection*{Funding Statement}
\noindent{The Agriculture and Food Research Initiative} (AFRI) grant no. 2020-67021-32799/project accession no.1024178 from the USDA National Institute of Food and Agriculture. 

\noindent{U.S. Department of Energy}, Office of Science, Biological and Environmental Research Program under Award Number DE-SC0018420

\section*{Acknowledgments}
This work is supported by (1) The Agriculture and Food Research Initiative (AFRI) grant no. 2020-67021-32799/project accession no.1024178 from the USDA National Institute of Food and Agriculture. The views and conclusions are those of the authors and should not be interpreted as representing the official policies of the funding agencies or the government. (2) The DOE Center for Advanced Bioenergy and Bioproducts Innovation (U.S. Department of Energy, Office of Science, Biological and Environmental Research Program under Award Number DE-SC0018420). Any opinions, findings, and conclusions or recommendations expressed in this publication are those of the author(s) and do not necessarily reflect the views of the U.S. Department of Energy.

\bibliographystyle{IEEEtranN}
\bibliography{reference.bib}

\clearpage

% \textbf{\LARGE Appendix}
\twocolumn[
\begin{center}
\vspace{2mm}
\hrule height 0.6pt
\vspace{2mm}
{\LARGE \textbf{Supplementary Information}}
\vspace{2mm}
\hrule height 0.6pt
\vspace{2mm}
\end{center}
]

\setcounter{equation}{0}

\section*{Technical Terms}
\label{ap: glossary}
To enhance readability for a broader audience, this section provides intuitive explanations of key technical concepts underlying flow-based generative models. We start from basic geometric and probabilistic notions, then move toward dynamic systems and learning objectives. To help readers from diverse scientific backgrounds, we group key technical concepts into three categories: 
(1) \textbf{Mathematical foundations}, which describe the geometric and dynamical principles underlying generative modeling; 
(2) \textbf{Machine learning concepts}, which capture probabilistic and representational ideas; and 
(3) \textbf{Generative Modeling paradigms}, which combine these principles into complete generative frameworks.

\subsection*{Mathematical Foundations}

\textbf{Manifold} (\url{https://en.wikipedia.org/wiki/Manifold}).
A manifold is a smooth surface (possibly curved) embedded in a higher-dimensional space. It generalizes familiar shapes such as a line (1D) or a sphere (2D) to any dimension. 
% In biological data, although measurements such as gene expression or protein conformations live in a very high-dimensional space, their true variability often lies along a low-dimensional manifold. For instance, a developmental trajectory of cells can be thought of as a continuous curve (1D manifold) through gene-expression space.

\textbf{Simplex} (\url{https://en.wikipedia.org/wiki/Simplex}).
A simplex is a geometric object representing proportions or compositional data. A 2D simplex is a triangle where every point represents a mixture of three components that sum to one (e.g., cell type proportions, nucleotide compositions). Many biological quantities are naturally constrained to a simplex, which is why modeling compositional data requires respecting this geometric structure.

% \paragraph{Probability Distribution.}
% A probability distribution describes uncertainty about where data points occur. For instance, the distribution of gene expression across cells or conformations of a protein. Generative models learn to represent such distributions — ideally, they should be able to sample new points from it that look realistic.

\textbf{Vector Field} (\url{https://en.wikipedia.org/wiki/Vector_field}).
A vector field assigns a direction and magnitude of change to every point in space. It describes ``how data moves''. In flow-based modeling, the vector field governs how a sample evolves continuously from a simple distribution (like Gaussian noise) to a complex one (like a real biological state). This field is the central object learned by flow matching methods.

\textbf{Ordinary Differential Equation (ODE)} (\url{https://en.wikipedia.org/wiki/Ordinary_differential_equation}).
An ODE specifies how a quantity \( x(t) \) changes continuously over time \( t \) according to a rule
\begin{equation}
\frac{dx}{dt} = v(x,t),
\end{equation}
where \( v(x,t) \) is called the \emph{vector field} and determines the instantaneous direction and speed of change. 
Here, the unknown is the trajectory \( x(t) \): how the state \( x \) evolves as time progresses. 
Solving the ODE means finding this function \( x(t) \) given an initial condition \( x(0) \), which traces a continuous path through space following the vector field.

\subsection*{Machine Learning Concepts}

\textbf{Latent Space} (\url{https://en.wikipedia.org/wiki/Latent_space}).
Latent space refers to a compressed, abstract representation learned by models. For example, cells with similar functional states or proteins with similar structures may occupy nearby regions in latent space. Generative models often operate in this space to simplify learning and sampling while preserving essential structure.

\textbf{Score Function} (\url{https://en.wikipedia.org/wiki/Informant_(statistics)}).
Let \( p(x) \) denote the probability density function that describes how likely each data point \( x \) is to occur. 
The \emph{score function} is defined as the gradient of the log-probability density with respect to \( x \), written as \( \nabla_x \log p(x) \). 
It points in the direction where the probability increases most rapidly, indicating how to move a sample toward regions of higher likelihood. 
In diffusion or score-based generative models, learning this function allows the model to iteratively “denoise” samples, pushing them toward realistic biological configurations. 
It plays a similar role to the vector field but focuses on local probability gradients rather than explicit trajectories through time.

\textbf{Kullback–Leibler Divergence} (\url{https://en.wikipedia.org/wiki/Kullback-Leibler_divergence}).
The Kullback–Leibler (KL) divergence measures how one probability distribution \( q(x) \) differs from another reference distribution \( p(x) \). 
Formally,
\begin{equation}
D_{\mathrm{KL}}(q \| p) = \int q(x) \, \log \frac{q(x)}{p(x)} \, dx.
\end{equation}
Intuitively, \( D_{\mathrm{KL}}(q \| p) \) quantifies the extra information (in bits or nats) required to describe samples from \( q \) when using a code optimized for \( p \). 
In generative modeling, \( q(x) \) often represents the model's learned distribution and \( p(x) \) the true data distribution. 
Minimizing KL divergence encourages the model to generate samples that are statistically indistinguishable from real data.

\subsection*{Generative Modeling Paradigms}

\textbf{Normalizing Flow} (\url{https://en.wikipedia.org/wiki/Flow-based_generative_model}).
A normalizing flow is a generative model that constructs a complex target distribution by transforming a simple, known base distribution (such as a standard Gaussian) through \textbf{a sequence of invertible and differentiable mappings}. 
Each transformation slightly warps the data space, and together these mappings produce a highly flexible distribution capable of capturing complex data structures. 
Because all transformations are invertible, both density evaluation and sampling remain efficient. 
Further details on normalizing flows are provided in Section "Flow-Based Model".

\textbf{Diffusion Model} (\url{https://en.wikipedia.org/wiki/Diffusion_model}).
A diffusion model is a generative framework that learns to \textbf{reverse a gradual noising process}. 
It starts from real data and progressively adds stochastic noise through a forward process, transforming the data distribution into a simple Gaussian. 
The model is then trained to learn the reverse process, i.e., how to remove noise step by step to recover realistic samples from pure noise.

\textbf{Flow Matching and its Advantages.}
Flow matching is a generative modeling paradigm that \textbf{learns the vector field} that transports one distribution into another. 
Unlike normalizing flows, which require each mapping to be exactly invertible and rely on computing Jacobian determinants for likelihood evaluation, flow matching learns the underlying dynamics of this transformation directly. 
Moreover, in contrast to diffusion models, FM avoids stochastic perturbations and instead models deterministic trajectories governed by an ordinary differential equation (ODE).

\section*{Mathematical Formulations and Guidance}
This section serves as a reference sheet for readers seeking rigorous formulations of the major tasks covered in this review and a deeper understanding of the role flow matching plays in each. We outline how these tasks are mathematically posed, summarize key derivations, and trace the evolution of the underlying formulations. Building on the flow-matching fundamentals introduced in Section "Flow-Matching Basics", we highlight how recent works instantiate and extend these principles across different domains. For complete technical accuracy, we encourage readers to consult the original papers, which provide additional details.

\subsection*{Sequence Generation}
\setcounter{paragraph}{0}

\paragraph{DNA sequence}
each DNA sequence $x$ of length $L$ over alphabet $\mathcal{A}$ is represented as a point on the product simplex, for example via position-wise categorical probability vectors $z = (z^{(1)},\dots,z^{(L)})$ with $z^{(\ell)} \in \Delta^{|\mathcal{A}|-1}$. 
A generic discrete flow-matching formulation defines a probability path $q_t(\cdot \mid x_0)$ on this simplex that connects a simple base distribution $q_1$ (e.g., uniform over $\Delta$) to the empirical distribution concentrated near the one-hot encoding of $x_0$ at $t=0$. The learnable velocity field $v_\theta(z,t)$ then minimizes the conditional flow matching loss
\begin{equation}
% \small
  \mathcal{L}_{\mathrm{DNA}}(\theta)
  \!=\! \mathbb{E}_{\substack{x_0 \sim \pi_0\\ t \sim \mathcal{U}(0,1)\\ z_t \sim q_t(\cdot \mid x_0)}}\! \,
    \Big[ \, \big\| v_\theta(z_t, t) - u_t(z_t \mid x_0) \big\|^2 \Big]
  \label{eq:dna-fm}
\end{equation}
where $u_t$ is the target velocity associated with the path $q_t$. 
% Sampling proceeds by integrating $\mathrm{d}z_t/\mathrm{d}t = v_\theta(z_t,t)$ from $t=1$ to $t=0$ starting at $z_1 \sim q_1$, and discretizing the final $z_0$ to obtain a DNA sequence. 
On top of the generic objective in \eqref{eq:dna-fm}, existing DNA flow-matching models mainly differ in how they parameterize probability paths and guidance on the categorical simplex. Fisher and Dirichlet flows \cite{davis2024fisher,stark2024dirichlet} endow the simplex with a non-Euclidean geometry via Fisher--Rao geodesics or mixtures of Dirichlet distributions, 
% , yielding smooth paths $q_t$ with closed-form velocities $u_t$ that support classifier(-free) guidance
whereas Gumbel-Softmax Flow Matching \cite{tang2025gumbel} operates in a relaxed logit space that scale well to high-dimensional sequences. Building on these base flows, subsequent work \cite{tang2025gumbel,DBLP:journals/corr/abs-2505-07086}
% views controllable and multi-objective DNA design as guidance on the learned dynamics, 
modify the sampling-time velocity $v_\theta$ using classifier gradients or rank-based multi-objective directions without retraining the underlying flow.

\paragraph{RNA sequence and structure generation.}
% RNA design extends the basic sequence setting by introducing rich conditioning and, in many cases, continuous structural degrees of freedom. 
Let $x_{\mathrm{seq}}$ denote an RNA sequence and $x_{\mathrm{str}}$ denote a 2D or 3D structure representation.
% (e.g., base-pair graph, backbone coordinates). 
A generic flow-matching formulation considers a state $z_t$ in a joint space $\mathcal{X}_{\mathrm{seq}} \times \mathcal{X}_{\mathrm{str}}$ and a condition $c$ encoding task-specific information.
% (secondary structure constraints, protein partners, family labels, etc.). 
The conditional probability path $q_t(\cdot \mid x_0, c)$ evolves from a simple base $q_1(\cdot \mid c)$ to the empirical joint distribution of $(x_{\mathrm{seq}},x_{\mathrm{str}})$ at $t=0$, and the conditional vector field $v_\theta$ is trained via
\begin{equation}
  \mathcal{L}_{\mathrm{RNA}}(\theta)
  \!=\! \mathbb{E}_{\substack{(x_0,c) \sim \pi_0\\ t \sim \mathcal{U}(0,1)\\ z_t \sim q_t(\cdot \mid x_0, c)}}\! \,
    \big\| v_\theta(z_t, t, c) - u_t(z_t \mid x_0, c) \big\|^2
  \label{eq:rna-fm}
\end{equation}
% where $v_\theta$ is typically factorized across sequence and structure components and may be constrained to be SE(3)-equivariant in the structural coordinates. 
% The central modeling choices are how broadly $c$ is defined and how tightly the sequence and structure flows are coupled within \eqref{eq:rna-fm}.
On top of the generic formulation in \eqref{eq:rna-fm}, RNA methods mainly differ in how they choose the state space and inject structural constraints. RNACG \cite{gao2024rnacg} instantiates \eqref{eq:rna-fm} purely in sequence space and treats a wide range of design problems, e.g., 3D inverse-folding targets, secondary structures, family labels, and functional annotations, as conditioning variable $c$.

\paragraph{Whole-Genome}
In single-cell genomics, each observation is a high-dimensional count vector $x \in \mathbb{N}^G$ over $G$ genomic features (e.g., genes or peaks). 
% Flow-matching models for these data typically operate either on a normalized compositional representation (e.g., $\ell_1$-normalized counts) or on a latent continuous space coupled with discrete likelihoods. 
A generic formulation keeps the flow in a continuous space $z_t \in \mathcal{Z}$.
% while tying it to observed counts through either a bijective transform or an explicit observation model. 
The conditional path $q_t(\cdot \mid x_0,c)$ connects a base $q_1(\cdot \mid c)$ to a latent representation of $x_0$ at $t=0$, with $c$ encoding optional attributes such as cell type. The corresponding loss is
\begin{equation}
\resizebox{0.9\linewidth}{!}{$
  \mathcal{L}_{\mathrm{cell}}(\theta)
  = \mathbb{E}_{\substack{(x_0,c) \sim \pi_0\\ t \sim \mathcal{U}(0,1)\\ z_t \sim q_t(\cdot \mid x_0, c)}}\! \,
    \Big[ \, \big\| v_\theta(z_t, t, c) - u_t(z_t \mid x_0, c) \big\|^2 \Big]
    + \lambda \, \mathcal{L}_{\mathrm{disc}}(\theta)
$}
  \label{eq:cell-fm}
\end{equation}
% is often supplemented with a discrete reconstruction term $\mathcal{L}_{\mathrm{disc}}$ (e.g., negative binomial likelihood) to encourage the terminal distribution of $z_t$ to decode back to realistic counts.

For whole-genome single-cell data, CellFlow \cite{palma2024cellflow} instantiates \eqref{eq:cell-fm} in a latent continuous space coupled with a compositional discrete likelihood. 
% CFGen \cite{palmamulti} lets the condition $c$ collect attributes such as cell type, batch, and modality indicators, turning $q_t(\cdot \mid x_0,c)$ into a conditional path that supports multi-modal, multi-attribute generation and rare-cell augmentation, while CellFlow \cite{palma2024cellflow} corresponds to the single-modality, largely unconditional special case. 
% In contrast, 
GENOT \cite{klein2024genot} uses flow matching to realize stochastic maps between cell distributions, augmenting \eqref{eq:cell-fm} with an entropic Wasserstein term.

\paragraph{Antibody}
% Antibody design is naturally formulated in a continuous structural state space subject to rigid-body symmetries, optionally coupled with discrete sequence variables. 
Let $x_{\mathrm{ab}} \in \mathbb{R}^{3N}$ denote the 3D coordinates of an antibody, and let $x_{\mathrm{ag}}$ denote the antigen structure. A generic SE(3)-equivariant flow-matching formulation defines a state $z_t$ that may include both antibody and antigen coordinates
% , and learns an equivariant vector field $v_\theta$ such that
\begin{equation}
  \frac{\mathrm{d} z_t}{\mathrm{d} t} = v_\theta(z_t, t, c), \quad c \in \{x_{\mathrm{ag}}, \text{interface prior}, \dots \}
\label{eq:antibody_obj}
\end{equation}
with a conditional path $q_t(\cdot \mid x_0, c)$ that connects a simple base distribution over structures to the distribution of physically valid complexes at $t=0$. The training loss mirrors \eqref{eq:cfm_obj}, but the norm is taken in an SE(3)-equivariant feature space.
% and $v_\theta$ is constrained such that $v_\theta(R z_t, t, c) = R v_\theta(z_t, t, c)$ for all rigid motions $R$.

% For antibody design, 
% IgFlow and dyAb \cite{nagaraj2024igflow,tan2025dyab} instantiate \eqref{eq:cfm_obj} in an SE(3)-equivariant structural space, learning vector fields over 3D backbones rather than over sequences. 
Based on \eqref{eq:antibody_obj}, IgFlow \cite{nagaraj2024igflow} treats the unconditional case where $z_t$ contains only antibody variable-domain coordinates and $q_t$ noises backbones toward a base distribution, while dyAb \cite{tan2025dyab} extends the same formulation to antigen-conditioned design by including both antigen and antibody coordinates in $z_t$ and encoding an AlphaFold2-predicted pre-binding antigen in $c$.
% and combining coarse interface alignment with a refined SE(3)-equivariant flow for the complex.

\subsection*{Molecule Generation}
\setcounter{paragraph}{0}

% Flow matching plays a central role in modern molecular generative models, where the goal is to generate valid, diverse, and property-aware molecules while respecting the underlying symmetries of 3D space. 
In this subsection, we first introduce equivariant and discrete flow matching as the foundational formulation for molecular systems, then extend it to fully SE(3)-equivariant 3D molecule generation, and finally discuss guided and conditional variants that incorporate structural or property-based constraints relevant to practical molecular design.

\paragraph{Equivariant Flow Matching and Discrete Flow Matching}
Equivariance and discrete structure are the foundational properties for molecular graph generation. In the simplified 2-dimensional case, let a molecular graph be denoted $G$. The generative task is to transport a simple prior $p_0(G)$ to the molecular data distribution $p_1(G) = p_{\mathrm{data}}(G)$.

% In continuous flow matching, a probability path $\{p_t\}_{t\in[0,1]}$ is induced by the ODE
% \[
% \frac{d x_t}{d t} = v_\theta(x_t,t), \qquad x_0 \sim p_0,
% \]

\textit{Permutation Equivariance.}
Let $P$ be a permutation matrix acting on graphs, an equivariant molecular flow matching model satisfies distributional invariance in the form of
\begin{align}
p_t(P\!\cdot\! G) &= p_t(G)
% R_{\theta,t}(P\!\cdot\!G, P\!\cdot\!G') &= R_{\theta,t}(G,G'), 
\label{eq:equivariance}
\end{align}

% ============================================================
% CatFlow / VFM
% ============================================================
CatFlow achieves permutation equivariance based on Variantional Flow Matching \cite{eijkelboom2024variational}, which rewrites the marginal flow using the posterior over endpoints, introduces a variational approximation $q_{\theta,t}(x_1\mid x)$, and minimizes 
\begin{equation}
\label{eq:vfm}
\mathcal{L}_{\mathrm{VFM}}
=
\mathbb{E}_{t,\,x,x_1\sim p_t(x,x_1)}
\left[
-\log q_{\theta,t}(x_1\mid x)
\right]
\end{equation}

GGFlow \cite{DBLP:journals/corr/abs-2411-05676} defines a probability path using an \emph{optimal transport coupling}  
$\gamma(G_0,G_1)$ between $p_0$ and $p_1$, with permutation-invariant Hamming cost $\gamma$.
The model approximates $p(G_1\mid G_t)$ via a variational posterior $q_{\theta,t}(G_1\mid G_t)$ and minimizes a VFM-style objective:
\begin{equation}
\label{eq:ggflow}
\mathcal{L}_{\mathrm{GGFlow}}
=
\mathbb{E}_{t,G_t,G_1}
\left[
-\log q_{\theta,t}(G_1\mid G_t)
\right]
\end{equation}
GGFlow implements $q_{\theta,t}$ with a permutation-equivariant transformer (GraphEvo) to achieve permutation equivariance for the entire generative process.

\textit{Discrete Flow Matching.} 
Discrete flow matching extends flow-based generative modeling to finite, combinatorial state spaces, making it particularly well suited for molecular graphs, where both atom types and bond types are categorical variables.
DeFoG instantiates this idea by optimizing
\begin{equation}
\mathcal{L}_\text{DeFoG}=\mathbb{E}_{t \sim \mathcal{T}, p_1 (G_1), p_{t|1}(G_t| G_1)} \operatorname{CE}_\lambda (G_1, \mathbf{p}^{\theta}_{1|t} (\cdot|G_t))
\end{equation}
where $\operatorname{CE}_\lambda  (G_1, \mathbf{p}^{\theta}_{1|t} (\cdot|G_t))$ is:
\begin{equation}
    \scalebox{0.88}{$
         - \sum_n \operatorname{log} \left(p^{\theta,(n)}_{1|t} (x^{(n)}_1|G_t) \right) - \lambda \sum_{i < j} \operatorname{log} \left(p^{\theta,(ij)}_{1|t} (e^{(ij)}_1|G_t) \right)
    $}
\end{equation}
% Here, $\lambda \in \mathbb{R}^+$ is introduced to weight nodes and edges differently to more flexibly capture varying topologies.

\paragraph{3D Molecule Generation with SE(3)-equivariant}
To extend equivariant flow matching from 2D molecular graphs to full 3D molecular, the generative model must operate jointly over discrete atom/bond types and continuous atomic coordinates.
A 3D molecule is represented by
$G = (Z, E, X)$ 
where $Z \in \mathcal{X}^N$ are atom types, $E \in \mathcal{E}^{N\times N}$ are bond types, and 
$X = (x_1,\dots,x_N)\in\mathbb{R}^{N\times 3}$ contains atomic coordinates.
Similar to 2D case, the continuous component evolves under an ODE
\begin{equation}
\frac{dX_t}{dt} = v_\theta(X_t,Z_t,E_t,t)
\end{equation}
with $v_\theta$ satisfying equivariance:
\begin{equation}
\scalebox{0.90}{$
    v_\theta(RXP^\top + \mathbf{1}r^\top,\,PZ,\,PEP^\top,t)
    =
    R\,v_\theta(X,Z,E,t)\,P^\top
$}
\end{equation}

We discuss some representative instantiations of 3D equivariant flow matching.
EquiFlow \cite{DBLP:journals/corr/abs-2412-11082} focuses on conditional conformation generation $p(X\mid C)$ for a fixed molecular graph $C=(Z,E)$. 
It adopts the standard conditional CFM loss
\begin{equation}
\mathcal{L}_{\mathrm{CFM}}^{\mathrm{cond}}
=
\mathbb{E}
\left[
\|v_\theta(X_t,C,t)-u_t(X_t\mid X_1,C)\|^2
\right]
\end{equation}
but constructs $u_t$ using an OT path based on: center $X^0$, compute the Kabsch rotation $R^\star$, and interpolate between $X^0$ and $R^\star X_1$. 
This produces an SE(3)-invariant path tailored to conformation prediction, paired with an SE(3)-equivariant backbone.
Megalodon \cite{reidenbach2024applications} jointly trains
\begin{align}
X_t &= \alpha(t)\varepsilon + \beta(t) X_1 \\ %\,\,\,
\mathcal{L}_{\mathrm{CFM}}^{\mathrm{coord}}
&=
\|v_\theta(X_t)-(\dot\alpha(t)\varepsilon+\dot\beta(t)X_1)\|^2
\end{align}
% \textit{Builds on previous methods by:} scaling SE(3)-equivariant FM with a unified architecture for discrete and continuous variables, without introducing new OT formulations.
Additionally, Equivariant Variational Flow Matching \cite{equivariantvariationalfm} extends previous 2D VFM and CatFlow \cite{eijkelboom2024variational} for constraint-driven and symmetry-aware generation.

\paragraph{Guided and Conditional Molecule Generation}
Guided and conditional flow matching extend the basic formulation by incorporating external signals, such as preferences, structural priors, categorical inputs, or energy functions.
Given a conditioning variable $c$ or a guiding signal $g$, the goal is to learn
\begin{equation}
\frac{dX_t}{dt} = v_\theta(X_t, c, g, t)
\end{equation}
% \textbf{FlowDPO}~\cite{EpusiLXfNd} introduces preference-based guidance for 3D atomic structure prediction.  
% Starting from the standard CFM objective
% \[
% \mathcal{L}_{\mathrm{CFM}}
% =
% \mathbb{E}\bigl[\|v_\theta(X_t,t)-u_t(X_t\mid X_1)\|^2\bigr],
% \]
% FlowDPO augments training with a Direct Preference Optimization (DPO) term comparing preferred and disfavored structures,
% \[
% \mathcal{L}_{\mathrm{DPO}}
% = 
% -\log \sigma\!\left(
% \beta\bigl(\log p_\theta(x^+)-\log p_\theta(x^-)\bigr)
% \right),
% \]
% guiding the learned flow toward higher-quality conformations while reducing hallucinations.  
% This represents a shift from \emph{data-only} FM to \emph{preference-aligned} FM for 3D structures.
Based on this, Instead of assuming a standard velocity field satisfying $\partial_t p_t + \nabla\!\cdot(p_t v_t)=0$, Extended Flow Matching \cite{DBLP:journals/corr/abs-2402-18839} introduces a generalized transport operator $\mathcal{A}_t$
% \[
% \partial_t p_t = \mathcal{A}_t[p_t],
% \]
allowing richer inductive biases.
% The corresponding conditional objective becomes
% \[
% \mathcal{L}_{\mathrm{EFM}}
% =
% \mathbb{E}\bigl[\|v_\theta(X_t,c,t)-u^{\mathrm{gen}}_t(X_t\mid X_1,c)\|^2\bigr],
% \]
% where $u^{\mathrm{gen}}_t$ is derived from the generalized transport.  
% This extends FM to domains requiring more flexible conditional dynamics.
FlowMol ~\cite{dunn2024mixed} decomposes the state and pairs continuous CFM 
% \[
% \mathcal{L}_{\mathrm{CFM}}^{\mathrm{cont}}
% =
% \|v_\theta(X_t,c,t)-u_t(X_t\mid X_1,c)\|^2,
% \]
with a discrete flow-matching loss similar to DeFoG.  
Energy-Based Flow Matching \cite{zhouenergy2025} incorporates an explicit energy function $E(X)$ into both training and inference.  
% The reference velocity is modified as
% \[
% u_t^{\mathrm{EB}}(X_t\mid X_1)
% =
% u_t(X_t\mid X_1) 
% - \lambda\,\nabla_X E(X_t),
% \]
% and the training objective minimizes
% \[
% \mathcal{L}_{\mathrm{EBFM}}
% =
% \mathbb{E}\bigl[\|v_\theta(X_t,t)-u_t^{\mathrm{EB}}(X_t\mid X_1)\|^2\bigr],
% \]
% biasing the flow toward low-energy, physically plausible structures.  
Energy signals act as auxiliary guidance without requiring explicit conditional variables.
OC-Flow \cite{DBLP:journals/corr/abs-2410-18070} frames flow matching guidance as an optimal control problem.  
Given a pre-trained flow $v_\theta$, OC-Flow seeks an optimal control input $u_t$.
% solving
% \[
% \frac{dX_t}{dt} = v_\theta(X_t,t) + u_t,\qquad
% u_t^\star = \arg\min_{u_t}
% \!\int_0^1 \!\!\ell(X_t,g) + \tfrac{\lambda}{2}\|u_t\|^2 dt,
% \]
% where $\ell$ measures deviation from the target guidance signal $g$.  
This yields training-free guidance that preserves the original flow while steering samples toward desired geometric or biochemical constraints.

% Taken together, these variants extend the basic flow-matching formulation by incorporating preferences (FlowDPO), generalized transport operators (EFM), mixed discrete–continuous conditioning (FlowMol), physical energy landscapes (EBFM), and optimal-control-based inference-time steering (OC-Flow).  
% They highlight the increasing flexibility of flow matching for generating biologically constrained and chemically meaningful 3D molecular structures.

\subsection*{Protein Generation}
\setcounter{paragraph}{0}

Protein generation encompasses a diverse set of tasks, making a single unified problem definition challenging. Accordingly, in the mathematical guidance section, we present formulations on a task-by-task basis.

\paragraph{Backbone Generation}
We represent a protein backbone of length $N$ as a collection of rigid frames 
$X = (T^{(1)},\dots,T^{(N)}) \in \mathrm{SE}(3)^N$, where each residue frame 
$T^{(n)} = (r^{(n)}, x^{(n)})$ consists of a rotation 
$r^{(n)} \in \mathrm{SO}(3)$ and a translation $x^{(n)} \in \mathbb{R}^3$. 
The goal is to learn a time-dependent $\mathrm{SE}(3)^N$-equivariant vector 
field $v_\theta(t, X_t)$ that transports a simple prior distribution 
$\rho_0$ 
% (e.g., isotropic rotations and Gaussian translations) 
 to the empirical 
backbone distribution~$\rho_1$. 
% This transport is defined by
% flow ODE:
% \[
% \frac{dX_t}{dt} = v_\theta(t, X_t), \qquad 
% X_0 \sim \rho_0,\; X_1 \sim \rho_1,\; t \in [0,1].
% \]

Flow matching 
decomposes the unknown intermediate marginals 
$\rho_t$ into tractable conditional geodesic bridges between paired samples 
$(X_0, X_1)$. For each residue $n$, the conditional interpolation 
$T_t^{(n)} = (r_t^{(n)}, x_t^{(n)})$ follows:
% the geodesic on $\mathrm{SE}(3)$:
\begin{equation}
    x_t^{(n)} = (1-t)x_0^{(n)} + t x_1^{(n)} 
\end{equation}
\begin{equation}
r_t^{(n)} = 
\exp_{r_0^{(n)}}\!\left(t\,\log_{r_0^{(n)}}(r_1^{(n)})\right)
\end{equation}
where the translations are linearly interpolated and the rotations follow the 
shortest geodesic in $\mathrm{SO}(3)$. The corresponding conditional velocity 
fields admit closed forms:
\begin{equation}
\dot{x}_t^{(n)} = \frac{x_1^{(n)} - x_t^{(n)}}{1-t}
\qquad
\dot{r}_t^{(n)} = 
\frac{\log_{r_t^{(n)}}(r_1^{(n)})}{1-t}
\end{equation}
The flow-matching objective regresses the learned vector field to this 
ground-truth bridge:
\begin{equation}
\begin{aligned}
   \mathcal{L}_{\text{FM}}(\theta)
&=
\mathbb{E}_{t,(X_0,X_1)}
\!\left[
\sum_{n=1}^N
\Big(
\|v^{(n)}_x(t,X_t) - \dot{x}_t^{(n)}\|_2^2\right.\\
&\left.+
\|v^{(n)}_r(t,X_t) - \dot{r}_t^{(n)}\|_{\mathrm{SO}(3)}^2
\Big)
\right] 
\end{aligned}
\end{equation}
% often implemented by reparameterizing $v_\theta$ to predict the clean frame 
% $X_1$ from noisy states $X_t$ using an $\mathrm{SE}(3)$-equivariant backbone 
% transformer.

% Under this general formulation, many works design different variants for special goals. For example, \emph{FrameFlow} realizes deterministic 
% $\mathrm{SE}(3)$ flow matching with improved geodesic schedules, alternative 
% rotational priors, and pre-alignment strategies that stabilize learning and 
% reduce integration depth. 
% \emph{FoldFlow} adopts the same ODE framework but improves the conditional path 
% construction using Riemannian optimal transport couplings and, in its stochastic 
% variant, Brownian bridges on $\mathrm{SO}(3)$, providing smoother flows and 
% enhanced structural diversity.
% \emph{FoldFlow-2} further couples the flow on rotations and translations with 
% sequence information by learning a conditional vector field 
% $v_\theta(t, r_t^{(n)}, x_t^{(n)}, \bar{a})$ given masked sequence embeddings 
% $\bar{a}$. 
% The resulting conditional loss,
% \[
% \mathcal{L}_{\text{cond}}(\theta)
% =
% \mathbb{E}_{t,(X_0,X_1,\bar{a})}
% \!\left[
% \|v_\theta(t, r_t,\bar{a}) - \log_{r_t}(r_0)/t\|_{\mathrm{SO}(3)}^2
% +
% \|v_\theta(t, x_t,\bar{a}) - (x_t - x_0)/t\|_2^2
% \right],
% \]
% unifies unconditional backbone generation, sequence-conditioned folding, and 
% structure/sequence inpainting within a single flow-matching framework.

\paragraph{Co-design Generation}
We represent a protein of length $L$ as
\begin{equation}
x = (s, z), \qquad 
s \in \mathcal{A}^L,\quad 
z \in \mathcal{Z}^L
\end{equation}
where $s$ is the amino-acid sequence and $z$ is the structural representation.
Co-design aims to learn a generative model $p_\theta(x)$ that transports a simple base
distribution $p_0(x)$ to the data distribution $p_{\text{data}}(x)$.

Flow-based methods introduce a time-indexed path $(p_t)_{t\in[0,1]}$ between $p_0$ and $p_{\text{data}}$, defined via conditional reference paths $p_t(x_t \mid x_1)$ for $x_1 \sim p_{\text{data}}$ and $x_0 \sim p_0$.
In the continuous channel, an $\mathrm{SE}(3)$-equivariant interpolation $z_t=\psi_t(z_0,z_1)$ induces the target velocity
\begin{equation}
u_t^{\text{cont}}(x_t,x_1)=\partial_t z_t(z_0,z_1)
\end{equation}
In the discrete channel, sequence evolution is modeled by a continuous-time Markov chain (CTMC) with generator $Q_t(x_1)$, satisfying
\begin{equation}
\partial_t p_t^{\text{disc}}(\cdot\mid s_1)
=
Q_t(x_1)^\top p_t^{\text{disc}}(\cdot\mid s_1)
\end{equation}
Assuming a factorized reference path,
\begin{equation}
p_t(x_t\mid x_1)=
p_t^{\text{disc}}(s_t\mid s_1)\,
p_t^{\text{cont}}(z_t\mid z_1)
\end{equation}
the model parameterizes a joint field
\begin{equation}
v_\theta(x_t,t)=
\big(v_\theta^{\text{cont}}(x_t,t),\, Q_\theta(x_t,t)\big)
\end{equation}
yielding coupled dynamics: an ODE for $z_t$ and a CTMC for $s_t$.

Flow matching trains $v_\theta$ to align with the reference flow in expectation over $(t,x_t,x_1)$. A practical loss decomposes into continuous and discrete terms:
\begin{equation}
\resizebox{0.88\linewidth}{!}{$
\mathcal{L}_{\mathrm{FM}}(\theta)
=
\mathbb{E}\Big[
\|v_\theta^{\text{cont}} - u_t^{\text{cont}}\|_2^2
+
\lambda\,
\mathrm{CE}\big(
q_\theta(\cdot\mid x_t,t),
q^\star(\cdot\mid x_t,x_1,t)
\big)
\Big]
$}
\end{equation}
where $q_\theta$ and $q^\star$ are one-step jump distributions from $Q_\theta$ and $Q_t$.
Sampling draws $x_0\sim p_0$ and integrates the coupled flow to $t=1$, producing a coherent sequence--structure pair $(s_1,z_1)$.

% \paragraph{Instantiations in Protein Co-design.}
% Under this formulation, Campbell et al.\ instantiate the discrete channel with 
% Discrete Flow Models (DFMs), where the CTMC generator $Q_t$ is chosen such that the 
% interpolation between a token prior and the data reduces to closed-form dynamics 
% on the simplex, enabling a pure flow-matching objective in cross-entropy. Their 
% \emph{Multiflow} model combines this discrete flow with a continuous structural 
% flow (e.g., FrameFlow-style dynamics), allowing flexible conditional sampling such 
% as predicting sequence from structure or structure from sequence.

% Yang et al.\ instead discretize both modalities by encoding 3D backbones into 
% structure tokens via a VQ-VAE and learning a joint discrete flow over $(s,r)$ 
% in the shared token space. Their \emph{CoFlow} model implements the joint path 
% $p_t(x_t \mid x_1)$ using linear token interpolation and balanced masking, and 
% parameterizes $Q_\theta$ with a bidirectional transformer augmented with Fourier 
% time embeddings. Compared to the abstract formulation above, Multiflow realizes a 
% mixed continuous--discrete flow that preserves geometric fidelity, whereas CoFlow 
% specializes to a fully discrete joint flow that leverages powerful protein 
% language-model priors to scale co-design to long sequences and diverse generation tasks.

\paragraph{Motif-Scaffolding Generation}
Given a fixed functional motif $M$, i.e., a set of backbone frames or atomic coordinates that must be preserved, the goal of motif-scaffolding is to sample full protein backbones $X = (X_M, X_S)$ whose scaffold region $X_S$ is geometrically compatible with the prescribed motif $X_M = M$. Let the backbone lie in the product manifold $\mathcal{X} = \mathrm{SE}(3)^N$, and denote by $p_{\mathrm{data}}(X)$ the distribution over native backbones. To model $p(X_S \mid X_M=M)$, we construct conditional probability paths
\begin{equation}
\resizebox{0.85\linewidth}{!}{$
    X_t = \exp_{X_0}\!\big(t \log_{X_0}(X_1)\big),\quad 
X_0 \sim p_0,\;\; X_1 \sim p_{\mathrm{data}}(\cdot\mid M)
$}
\end{equation}
% where $p_0$ is an easy-to-sample prior on $\mathcal{X}$ and the exponential/logarithmic maps follow the $\mathrm{SE}(3)$ Riemannian geometry. 
The conditional vector field along this interpolant is
\begin{equation}
u(X_t,t\mid X_1) 
= \frac{1}{1-t}\,\log_{X_t}(X_1),
\end{equation}
and a neural network $v_\theta$ is trained to regress $u$ using the conditional flow-matching objective
\begin{equation}
\resizebox{0.88\linewidth}{!}{$
\mathcal{L}_{\mathrm{CFM}}
= \mathbb{E}_{t\sim U(0,1),\,X_1\sim p_{\mathrm{data}},\,X_0\sim p_0}
\Big[
\|u(X_t,t\mid X_1) - v_\theta(X_t,t,M)\|^2
\Big]
$}
\end{equation}
\paragraph{Pocket \& Binder Design}
We represent a protein--ligand complex as $C = (R,G)$, 
where the pocket $R$ consists of residues with types, backbone frames $(x^{(i)},O^{(i)})$, torsions $\chi^{(i)}$, and interaction labels, and the ligand $G$ is a molecular graph with 3D coordinates. 
% The state space $\mathcal{M}$ is a product manifold combining Euclidean coordinates, $\mathrm{SO}(3)$ rotations, torsion tori, and discrete residue/interaction labels.  
Given conditioning information $S$,
% \[
% S = (P \setminus R,\; \text{ligand 2D graph}),
% \]
% the goal is to learn a conditional distribution
% $
% p_\theta(C \mid S) \approx p_{\text{data}}(C \mid S)
% $ 
% over ligand-compatible pockets and binders, while respecting $\mathrm{SE}(3)$ symmetry.
Flow matching formulates this as a conditional flow $(\psi_t)_{t\in[0,1]}$ on $\mathcal{M}$ driven by an $\mathrm{SE}(3)$-equivariant vector field:
\begin{equation}
\frac{d}{dt} C_t = u_t(C_t \mid S), 
\,\,
C_0 \sim p_0(\cdot \mid S),\; 
C_1 \sim p_{\text{data}}(\cdot \mid S)
\end{equation}
For each $(C_0,C_1)$, we define a conditional path $p_t(\cdot \mid C_1,S)$ using geodesic interpolation on each manifold component, 
% (linear for coordinates, geodesics on $\mathrm{SO}(3)$ and torsions, relaxed paths for labels), 
yielding a reference field $u_t(C_t \mid C_1,S)$.
A neural $\mathrm{SE}(3)$-equivariant field $v_\theta(C_t,t \mid S)$ is trained via conditional flow matching:
\begin{equation}
\resizebox{0.88\linewidth}{!}{$
\mathcal{L}_{\mathrm{CFM}}(\theta)
=
\mathbb{E}_{S,\, C_1,\, t,\, C_t}
\Big[
\| v_\theta(C_t,t \mid S) - u_t(C_t \mid C_1,S) \|_g^2
\Big]
$}
\end{equation}
where $\|\cdot\|_g$ is the Riemannian metric on $\mathcal{M}$.

\paragraph{Conformer Prediction}
Proteins occupy ensembles of conformational states $X$, shaped by the amino acid sequence $s$ and often by experimental observables $y$ (e.g., energies, B-factors, NMR restraints). We model states on the product manifold
$
\mathcal{X} = \mathrm{SE}(3)^N \times \mathbb{T}^K,
$
where $\mathrm{SE}(3)^N$ encodes residue frames and $\mathbb{T}^K$ encodes side-chain torsions.  
Given a sequence $s$, the goal is to generate samples from the conditional ensemble $p_{\mathrm{data}}(X \mid s,y)$.
Flow matching learns a time-dependent vector field $v_\theta$ that transports a simple prior $p_0(X_0 \mid s,y)$ to the target distribution, typically with a conditional flow-matching (CFM) loss on $\mathrm{SE}(3)$

\paragraph{Side-chain Packing}
Given a fixed backbone $B$ and sequence $s$, side-chain packing aims to generate physically valid torsions $\chi=(\chi^{(1)},\dots,\chi^{(N)})$ that satisfy steric and energetic constraints. Each residue $i$ has $k_i\in\{0,\dots,4\}$ rotatable angles, so the full side-chain space is the torus
$\mathcal{X}=\mathbb{T}^K, K=\sum_i k_i$.
Flow matching offers a natural way to model the conditional distribution $p(\chi\mid B,s)$ on this manifold.

Let $p_{\mathrm{data}}(\chi \mid B,s)$ be the empirical distribution, and let $p_0(\chi_0)$ be a simple prior. For each pair $(\chi_0,\chi_1)$, we define a geodesic interpolation on the torus:
\begin{equation}
\chi_t=\exp_{\chi_0}\!\big(t\,\log_{\chi_0}(\chi_1)\big),
\quad 
\chi_1\sim p_{\mathrm{data}}(\cdot\mid B,s)
\end{equation}
% with coordinate-wise wrapped angle operations:
% \[
% \log_{\chi_0}(\chi_1)
% =\operatorname{atan2}(\sin(\chi_1-\chi_0),\cos(\chi_1-\chi_0)),
% \quad
% \exp_{\chi_0}(v)=(\chi_0+v)\bmod 2\pi.
% \]
This induces a reference vector field
\begin{equation}
u(\chi_t,t\mid\chi_1)=\tfrac{1}{1-t}\,\log_{\chi_t}(\chi_1)
\end{equation}
and $v_\theta$ is trained to match $u$ via the CFM loss.

\paragraph{Docking Prediction}
We model protein--ligand docking as learning a continuous transport from unbound (apo) to bound (holo) complexes. A complex is represented by heavy-atom coordinates
$
x = (x^{P}, x^{L}) \in \mathbb{R}^{3N_P} \times \mathbb{R}^{3N_L}
$
where $x^{P}$ and $x^{L}$ are protein and ligand atoms. For each training pair, we observe an apo conformation $x_0$, a holo conformation $x_1$, and conditioning information $c$. The goal is to learn
$
p_\theta(x_1 \mid x_0,c) \approx p_{\mathrm{data}}(x_1 \mid x_0,c)
$
and an approximately invertible map that transforms apo structures into their holo states.

Conditional flow matching enables simulation-free training by assigning a simple path for each $(x_0,x_1)$:
\begin{equation}
x_t = (1-t)x_0 + t x_1,\qquad t\in[0,1]
\end{equation}
and defining a target vector field 
$
u_t(x_t \mid x_0,x_1) = x_1 - x_0.
$
A neural field $v_\theta(x_t,t \mid c)$ is then trained to regress $u_t$ using the CFM loss.

% \begin{align}
% \mathcal{L}_{\mathrm{CFM}}(\theta)
% &= \mathbb{E}_{(x_0,x_1,c) \sim p_{\text{data}}} \;
%    \mathbb{E}_{t \sim \mathcal{U}[0,1]}
%    \Big[
%      \big\|
%        v_\theta(x_t, t \mid c) - u_t(x_t \mid x_0, x_1)
%      \big\|_2^2
%    \Big] 
%    \\
% &= \mathbb{E}_{(x_0,x_1,c), t}
%    \Big[
%      \big\|
%        v_\theta\big((1-t)x_0 + t x_1, t \mid c\big) - (x_1 - x_0)
%      \big\|_2^2
%    \Big].
% \end{align}
% In practice, many docking FM models further exploit the CondOT equivalence and reparameterize $v_\theta$ to directly regress the terminal state $x_1$ along the path, yielding an equivalent objective of the form
% \[
% \mathcal{L}_{\mathrm{CFM}}(\theta)
% = \mathbb{E}_{(x_0,x_1,c), t}
%    \Big[
%      \big\|
%        v_\theta(x_t, t \mid c) - x_1
%      \big\|_2^2
%    \Big],
% \]
% whose gradients match those of the full flow-matching objective. At inference time, docking is performed by integrating
% \[
% \frac{d}{dt} x_t = v_\theta(x_t, t \mid c),\qquad x_0 \sim p_0(\cdot \mid c),
% \]
% with a small number of ODE steps, yielding the predicted holo complex without any diffusion-time sampling or external energy minimization. When desired, SE(3) symmetry can be enforced by requiring $v_\theta$ to be equivariant under global rigid motions of the complex and the conditioning.

\paragraph{Peptide and Antibody Generation}
Peptide design can be framed as conditional flow matching over mixed geometric and categorical manifolds. Let
$
x=\{(a_j,R_j,x_j,\chi_j)\}_{j=1}^n
$
denote a peptide or antibody chain, where $a_j\in\Delta^{19}$ denotes a distribution over the 19 canonical amino-acid types, $(R_j,x_j)\in SE(3)$ is the backbone frame, and $\chi_j\in\mathbb{T}^k$ are torsions. Given conditioning information $c$ such as receptor geometry or epitope context, 
we learn a time-dependent vector field $v_\theta(x_t,t,c)$ such that
\begin{equation}
\frac{d}{dt}x_t=v_\theta(x_t,t,c),\qquad x_0\sim p_0
\end{equation}

For each pair $(x_0,x_1)$, CFM defines analytic interpolants on each manifold component: linear in $\mathbb{R}^3$, geodesic on $SO(3)$, wrapped-linear on $\mathbb{T}^k$, and simplex or smoothed discrete interpolation for residue identities.

\vfill

\end{document}